%% file: main.tex
\documentclass[sigconf]{acmart}
\usepackage{microtype}
\usepackage{graphicx}
\usepackage{subfigure}
\usepackage{multirow}
\usepackage{array}
\usepackage{booktabs}
\usepackage{subcaption}
\usepackage{xcolor}
\usepackage{makecell}
\usepackage{balance}
\usepackage{enumitem}
\usepackage{hanging}
\usepackage{hyperref}

\usepackage{makecell}
\usepackage{amsmath}

\usepackage{amssymb}
\usepackage{mathtools}
\usepackage{amsthm}
\usepackage{multirow}
\usepackage{arydshln}
\usepackage{pifont}
\usepackage{placeins}

\usepackage{algorithm}
\usepackage{algorithmic}

\definecolor{lightblue}{rgb}{0.21, 0.49, 0.74}

\usepackage[capitalize,noabbrev]{cleveref}

\newlength\savewidth\newcommand\shline{\noalign{\global\savewidth\arrayrulewidth
  \global\arrayrulewidth 1pt}\hline\noalign{\global\arrayrulewidth\savewidth}}
  
\newcommand{\cmark}{\ding{51}}%
\newcommand{\xmark}{\ding{55}}%

\AtBeginDocument{%
  }

\copyrightyear{2025}
\acmYear{2025}
\setcopyright{acmlicensed}\acmConference[MM '25]{Proceedings of the 33rd ACM International Conference on Multimedia}{October 27--31, 2025}{Dublin, Ireland}
\acmBooktitle{Proceedings of the 33rd ACM International Conference on Multimedia (MM '25), October 27--31, 2025, Dublin, Ireland}
\acmDOI{10.1145/3746027.3755110}
\acmISBN{979-8-4007-2035-2/2025/10}

\begin{document}

\title{StitchFusion: Weaving Any Visual Modalities to Enhance Multimodal Semantic Segmentation}

\settopmatter{authorsperrow=3}

\author{Bingyu Li}
\authornote{Work done during an internship at TeleAI}
\affiliation{%
  \institution{University of Science and Technology of China}
  \city{Hefei}
  \country{China}
}
\affiliation{%
  \institution{Institute of Artificial Intelligence (TeleAI), China Telecom}
  \city{Beijing}
  \country{China}
}
\email{libingyu0205@mail.ustc.edu.cn}

\author{Da Zhang}
\affiliation{%
  \institution{Northwestern Polytechnical University}
  \city{Xi’an}
  \country{China}
}
\affiliation{%
  \institution{Institute of Artificial Intelligence (TeleAI), China Telecom}
  \city{Beijing}
  \country{China}
}
\email{dazhang@mail.nwpu.edu.cn}

\author{Zhiyuan Zhao}
\affiliation{%
  \institution{Institute of Artificial Intelligence (TeleAI), China Telecom}
  \city{Beijing}
  \country{China}
}
\email{tuzixini@gmail.com }

\author{Junyu Gao}
\affiliation{%
  \institution{Institute of Artificial Intelligence (TeleAI), China Telecom}
  \city{Beijing}
  \country{China}
}
\email{gjy3035@gmail.com}

\author{Xuelong Li}
\authornote{Corresponding Author}
\affiliation{%
  \institution{Institute of Artificial Intelligence (TeleAI), China Telecom}
  \city{Beijing}
  \country{China}
}
\email{xuelong_li@ieee.org}

\renewcommand{\shortauthors}{Bingyu Li et al.}

\begin{abstract}
Multimodal semantic segmentation shows significant potential for enhancing segmentation accuracy in complex scenes. However, current methods often incorporate specialized feature fusion modules tailored to specific modalities, thereby restricting input flexibility and increasing the number of training parameters. To address these challenges, we propose \textit{StitchFusion}, a straightforward yet effective modal fusion framework that integrates large-scale pre-trained models directly as encoders and feature fusers. This approach facilitates comprehensive multi-modal and multi-scale feature fusion, accommodating any visual modal inputs.
Specifically, our framework achieves modal integration during encoding by sharing multi-modal visual information. To enhance information exchange across modalities, we introduce a multi-directional Modality Adapter module (\textit{MoA}) to enable cross-modal information transfer during encoding. By leveraging \textit{MoA} to propagate multi-scale information across pre-trained encoders during the encoding process, \textit{StitchFusion} achieves multi-modal visual information integration during encoding. Extensive comparative experiments demonstrate that our model achieves state-of-the-art performance on four multi-modal segmentation datasets with minimal additional parameters. Furthermore, the experimental integration of \textit{MoA} with existing Feature Fusion Modules (FFMs) highlights their complementary nature. Our anonymous code is \href{https://github.com/LiBingyu01/StitchFusion}{\textcolor{blue}{here}}. 
\end{abstract}

\ccsdesc[500]{Computing methodologies~Image segmentation}
\ccsdesc[500]{Computing methodologies~Image representations}
\ccsdesc[500]{Computing methodologies~Neural networks}

\keywords{Multimodal Semantic Segmentation, Modality Fusion, Modality Adapter}

\maketitle

\input{section/01_introduction}
\input{section/02_related_works}
\input{section/03_methods}
\input{section/04_experiments}
\input{section/05_conclusion}

\begin{acks}
This work was supported in part by the Natural Science Foundation of China under Grant 62306241, and in part by grants from the Innovation Foundation for Doctor Dissertation of Northwestern Polytechnical University (No.CX2025109).
\end{acks}


\bibliographystyle{ACM-Reference-Format}
\balance
\bibliography{ref}
%

\input{section/06_appendix}

\end{document}

%% file: section/01_introduction.tex
\section{Introduction}
\label{introduction}
Semantic segmentation is a critical vision processing technique extensively applied in scene understanding, change detection, and autonomous driving \cite{chen2017rethinking}. This technique enables systems to accurately recognize and interpret the surrounding environment by analyzing the semantic information of each image pixel \cite{chen2018encoder}. Despite the success of the models, most related research has primarily focused on single visual features (RGB), limiting their effectiveness in complex scenes and special environments, such as nighttime.

As application demands increase, processing a single modality is no longer sufficient to meet the requirements \cite{chen2017deeplab}. Consequently, the community is increasingly focusing on multimodal semantic segmentation\cite{joze2020mmtm}. Unlike traditional segmentation tasks that rely solely on RGB modalities, multimodal semantic segmentation leverages the complementary features of multiple visual modalities. Each modality can provide unique information: RGB captures color and texture, TIR captures thermal properties useful in low-light conditions, and depth sensors provide spatial and structural information \cite{zhang2023cmx}. By fusing these complementary modalities, segmentation models can achieve higher accuracy and robustness.

\begin{figure*}
\centering
\includegraphics[width=0.85\linewidth]{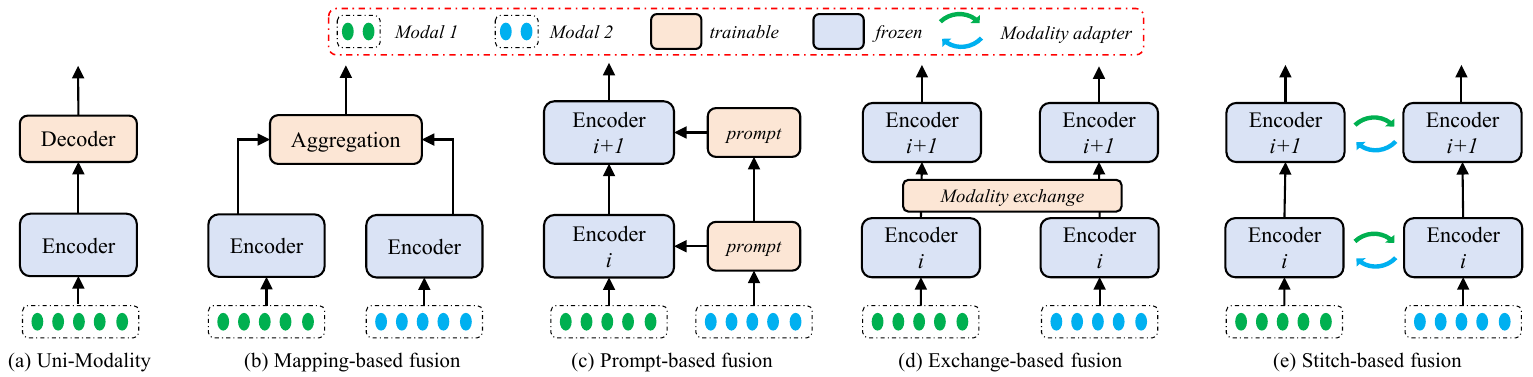} 
\vspace{-10pt}
\caption{Comparison of different model fusion paradigms. (a) Single modality processing. (b) Mapping-based fusion (c) Prompt-based fusion (d) Exchange-based fusion (e) Stitch-based fusion: feature fusion through direct information sharing with minimal parameters and no modality bias.}
\label{fig:fig_1}
\end{figure*}

Given the complementary nature of multimodal information, achieving the effective fusion of multiple visual modalities is crucial. 
While many innovative frameworks have been developed to enable modality fusion, efficient modality fusion remains an area that requires further exploration. As shown in \cref{fig:fig_1} (b-d), existing modality fusion models can be broadly categorized into three dominant paradigms: mapping-based methods\cite{zhang2023cmx,fu2019dual,zhou2023cacfnet}, prompt-based methods \cite{he2024prompting}, and exchange-based methods.

The mapping-based modality fusion approach maps multimodal features into a shared feature space, after which a modality fusion module combines these features\cite{kaykobad2023multimodal, li2024u3m}. Although this method can achieve certain fusion effects, the feature mappers and fusion modules often have a large number of parameters, sometimes even exceeding the parameter count of the encoder itself. To address this, the prompt-based approach introduces a more lightweight framework by incorporating additional modalities in the form of prompts on top of the RGB modality\cite{he2024prompting}. However, this method currently only supports the fusion of two modalities and introduces a degree of modality bias.
The exchange-based fusion framework has been explored recently, they remain limited to the fusion of two modalities \cite{wang2022multimodal} and suffer from high computational complexity \cite{jia2024geminifusion} during the modality exchange and fusion process.
Additionally, \cite{yin2023dformer} propose method for multimodal training from scratch. Although it is quite novel, it requires a substantial amount of training time and large-scale datasets. The number of trainable parameters is unacceptable in most cases. In this paper, we still apply pre-trained frozen backbones and take into account the limitations mentioned above in existing models. 

We reconsider the design requirements for modality fusion modules: (1) the fusion module should be efficient with a minimal number of additional parameters, (2) it should avoid introducing modality bias, and (3) it should be capable of adapting to arbitrary combinations of modalities for widely application. 

\begin{figure}
\centering
\includegraphics[width=\linewidth]{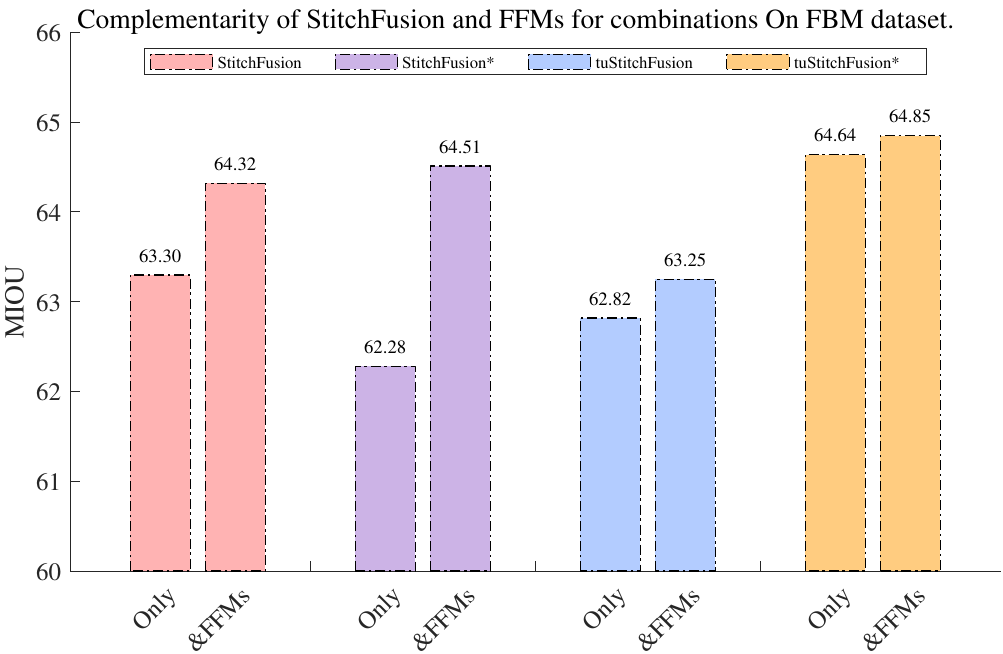}
\vspace{-12pt}
\caption{The StitchFusion framework (MiT-B3 as backbones) which can complement any other FFMs. We verify this using the FFM in \cite{kaykobad2023multimodal}. }
\vspace{-12pt}
\label{fig:fig_22222}
\end{figure}

To meet these criteria, we propose a framework called stitch-based, which achieves fusion by enabling interaction and sharing between modalities during the encoding stage. Unlike training from scratch, our framework directly adapts the backbone architecture with minimal parameters and enables training on smaller datasets instead of large-scale ones \cite{yin2023dformer}.
Inspired by these considerations, we propose a new stitch-based feature fusion model (\textit{StitchFusion}) that utilizes pretrained models as a feature fusion layer to demonstrate potential benefits. To achieve this target, we employ a simple Multidirectional Modality Adapter layer, named \textit{ModalityAdapter or MoA}, which shares and synchronizes modality-specific multi-scale information throughout the encoding process. This method leverages the encoder’s inherent multi-scale visual feature modeling capabilities, requiring fewer additional parameters for cross-modal fusion. Experimental results on \underline{7 standard datasets} and \underline{1 self-made underwater} multi-modal dataset (MMUS dataset in appendix) confirm that this new fusion paradigm not only outperforms traditional feature fusion methods but also enhances segmentation efficacy when combined (shown in Fig. \ref{fig:fig_22222}).

Our contributions are summarized as follows:

\begin{itemize}
    \item We introduce a multimodal feature fusion framework called \textit{StitchFusion}, which achieves cross-modal integration by enabling modality sharing during the encoding process. 
    \item We designed a Multidirectional Modality Adapter layer called MoA, which enables cross-modal information sharing. By this, our framework leverages the encoder's feature extraction capabilities to achieve effective feature fusion.
    \item Experiments on various multimodal semantic segmentation datasets with the StitchFusion for ViT-based encoder surpass previous state-of-the-art results. Comprehensive ablation studies suggest potential optimal placements and quantities for integrating the StitchFusion module to maximize modal fusion efficacy.
    \item The StitchFusion module and existing approaches based on additional FFMs can complement each other in terms of design and application, and we demonstrate their complementary nature through extensive experiments.
\end{itemize}

%% file: section/02_related_works.tex
\section{Related Work}
\label{related_work}
\subsection{Semantic Segmentation}
Semantic segmentation, a critical task in computer vision\cite{chen2024comkd}, has evolved significantly by developing various methods and models, particularly those leveraging convolutional neural networks (CNNs) and more recently, transformers \cite{xie2021segformer}. Early breakthroughs were achieved through fully convolutional networks (FCNs) \cite{long2015fully}, which enabled end-to-end pixel-wise predictions. Subsequent architectures, such as SegNet \cite{badrinarayanan2017segnet} and U-Net \cite{ronneberger2015u}, utilized encoder-decoder structures to capture both low-level and high-level features. The DeepLab series \cite{chen2017rethinking, chen2018encoder} introduced atrous convolutions and spatial pyramid pooling to enhance multi-scale context perception, while PSPNet \cite{zhao2017pyramid} aggregated context from different regions. Vision Transformer (ViT) models \cite{dosovitskiy2020image} leveraged self-attention mechanisms to capture long-range dependencies, with subsequent adaptations like SETR \cite{zheng2021rethinking} and Swin Transformer \cite{liu2021swin} improving computational efficiency and scalability. While single-modality data has seen substantial progress, multimodal semantic segmentation, integrating data such as RGB with other vision modalities, has been increasingly explored. The existing literature has proposed numerous feature fusion approaches \cite{hazirbas2016fusenet}. Building on these advancements, we propose the \textit{StitchFusion} model which introduces a novel feature fusion paradigm (\textit{StitchFusion}) using Multidirectional MLP layer (\textit{ Modality Adapter}) for effective multimodal integration.

\subsection{Vision Multimodal Fusion}
In classical visual tasks, single visual modalities often struggle to handle challenges in complex environments adaptively \cite{fu2019dual,zhang2023cmx}. Consequently, an increasing number of researchers are turning to multiple visual modalities, making the fusion of these modalities crucial. Some researchers have used fine-tuned pre-trained models to fuse multiple visual modalities, but this approach can lead to catastrophic forgetting. To mitigate this problem, many researchers have opted to freeze the pre-trained models, as seen in the work \cite{he2024prompting,cao2024bi} proposing a prompt-based method for multimodal fusion. Another paradigm utilizes multi-scale information and designs various FFMs to integrate the multi-scale information from each modality \cite{zhang2023delivering, dong2023egfnet, huang2019ccnet}. Although existing FFMs have achieved notable experimental results, they introduce excessive additional parameters and are often limited in the number of modality data they can handle. Therefore, in this paper, we introduce \textit{StitchFusion}, a simple but effective feature fusion framework incorporating the plug-and-play \textit{ Modality Adapter} module, which enables modality sharing with fewer parameters, thereby achieving feature fusion during the encoding process.

%% file: section/03_methods.tex
\section{Method}
\label{method}
This section introduces a new multimodal feature fusion framework for semantic segmentation, named \textit{StitchFusion}. As for how to achieve the feature sharing during the encoder stage, we use a simple multidirectional adapter.
\subsection{Preliminary and Analysis}
\subsubsection{Multimodality Segmentation Problem Definition}
Given an initial image containing different semantic pixel, the goal of RGB-based semantic segmentation is to learn a segmentation function $\mathcal{F}_{RGB}:\{ \mathbf{I}_{RGB}\} \rightarrow \mathbf{Y}$ that assigns a semantic label to each pixel in the image $\mathbf{I}_{RGB}$. For multi-modal semantic segmentation, given a set of multimodal images \( \mathbf{I} = \{\mathbf{I}_1, \dots, \mathbf{I}_m\} \), where \( \mathbf{I}_i \in \mathbb{R}^{H \times W \times C_i} \) represents data from the \(i\)-th modality, \( H \) and \( W \) are the height and width, and \( C_i \) is the number of channels. Different modalities provide distinct characteristics, such as RGB for color and texture, and depth maps for geometric details.
The goal is to predict pixel-level classifications through a model $\mathcal{F}_{MM}:\{\mathbf{I}_1, \mathbf{I}_2, \dots, \mathbf{I}_m\} \rightarrow \mathbf{Y} $, \( \mathbf{Y} \in \mathbb{R}^{H \times W \times L} \) is the predicted classification map, and \( L \) is the number of classes, and $\mathcal{F}_{MM}$ is the multi-modal segmentation model that integrates information from all modalities to produce a semantically segmented output.

\begin{figure*}
\centering
\includegraphics[width=0.88\linewidth]{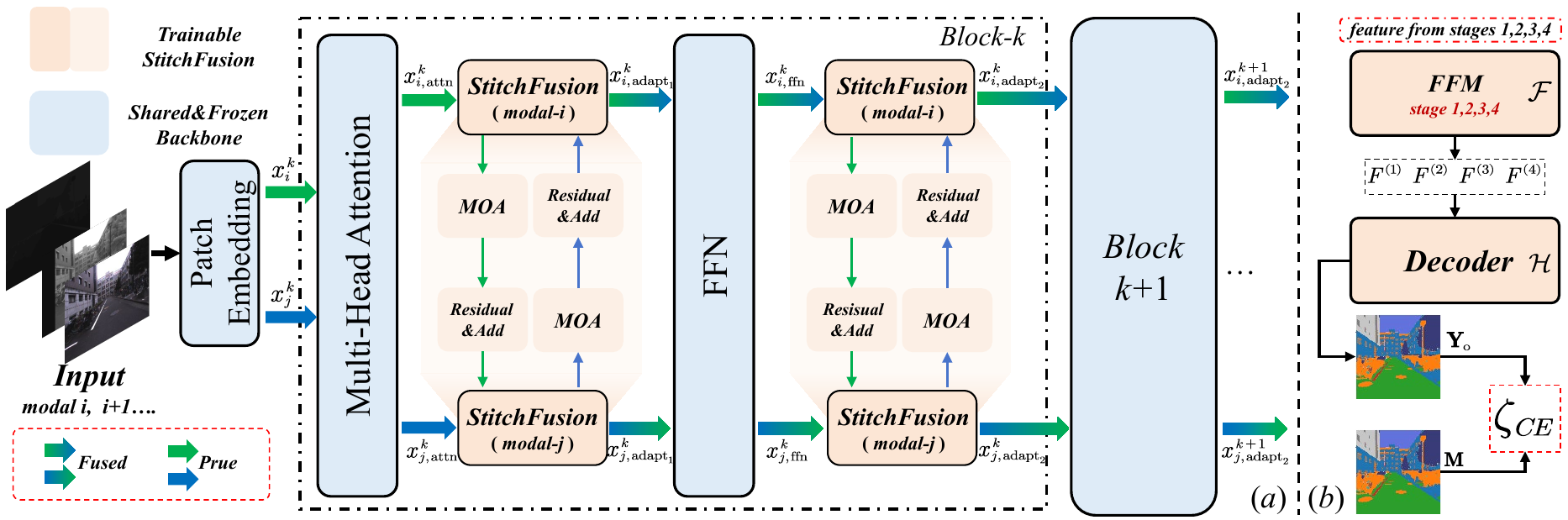}
\vspace{-12pt}
\caption{The overall StitchFusion framework. (a) Stitfusion Framework in Encoder Backbone, MOA denotes Modality Adapter. (b) Selective Feature Fusion Module(FFM) $\mathcal{F}$ and the MLP decoder $\mathcal{H}$.}
\label{fig:fig_3_}
\vspace{-10pt}
\end{figure*}

\subsubsection{Classic Feature Encoding and Fusion.}
The encoder progressively extracts features over four stages, capturing both local and global information. At each stage \( l \), the output for $i$-th modality image \( \mathbf{I}_i \) is \( Z_i^{(l)} \in \mathbb{R}^{H_l \times W_l \times C_l} \), where \( l = 1, 2, 3, 4 \) corresponds to different resolution levels. For the classic multimodal fusion strategy, features from the same stage across modalities are aggregated:
\begin{equation}
F^{(l)} = \mathcal{F}(Z_1^{(l)}, Z_2^{(l)}, \dots, Z_m^{(l)}),
\end{equation}
where \( \mathcal{F} \) could be concatenation, addition, or a more sophisticated interaction mechanism. This fusion occurs at each stage, producing multi-scale fused features \( F^{(1)}, F^{(2)}, F^{(3)}, F^{(4)} \). The fused features are then encoded into a lower-dimensional space for semantic segmentation:
\begin{equation}
\mathbf{Y}_{\text{o}} = \mathcal{H}([F^{(1)}, F^{(2)}, F^{(3)}, F^{(4)}]),
\label{equ_2}
\end{equation}
where \( \mathbf{Y}_{\text{o}} \) is the final output, and \( \mathcal{H} \) denotes the output head, it is a MLP as usual.

Although simple concatenation and summation operations can achieve modality fusion, their effectiveness is rather poor. Consequently, most classic paradigms are designed to be relatively complex \cite{zhang2023cmx, zhang2023delivering, kaykobad2023multimodal}. 
While adding a fusion head improves performance, it also increases the model complexity and parameters (we illustrate these framework in \cref{fig:fig_1}). Thus, we rethink the feature fusion strategy and design a simple yet efficient stitch-based feature fusion method, this framework is shown in \cref{fig:fig_1}(d).

\subsection{StitchFusion Framework.}
In this section, we introduce a novel stitch-based approach called StitchFusion, the illustration is shown in \cref{fig:fig_3_}. First, we present the feature encoding process, followed by a detailed description of the framework and the implementation of modality adapter.

\subsubsection{Feature Extraction And Encoding}

Given an input image of $i$-th modality $\mathbf{I}_i\in\mathbb{R}^{H \times W \times C_i}$, we use these model as the encoder to generate a feature map. To achieve information fusion, based on the ViT (both plain \cite{dosovitskiy2020vit} and swin  \cite{liu2021swin}) and Convnext architecture, we devise the StitchFusion \cite{liu2022convnet} framework by viewing the block as the encoder and feature fuser shown in \cref{fig:fig_3}.
As for the decoder, We use a simple MLP decoder as shown in Equ. \ref{equ_2}.

\subsubsection{StitchFusion for Vision Transformer Encoder}

To enable the pre-trained and frozen model to act as a modality fuser, we employ the Modality Adapter(\text{MoA}) Layer as an information stitcher within the encoder. Here, we use Vision Transformer (ViT) as an example for illustration.

First, the image $\mathbf{I}_i\in\mathbb{R}^{H \times W \times C_i}$ from the $i$-th modality is processed by the patch embedding module \(x_i=\text{PatchEmbed}(\mathbf{I}_i)\) to generate the input for the blocks.

Then for the $k$-th block, the attention mechanism processes the input feature map \( x_i^k \), corresponding to the \(i\)-th modality in the \(k\)-th block of the encoder. This step utilizes the multi-head self-attention (MHSA) mechanism to capture intra-modality dependencies and spatial relationships within the feature map \( x_i^k \):
\begin{equation}
x_{i,\text{attn}}^k = x_i^k + \text{DP}(\text{MHSA}(\text{LN}_1(x_i^k))),
\end{equation}
where \(\text{DP}\) denotes the DropPath operation, and \(\text{LN}_1\) represents Layer Normalization.

Next, the  Modality Adapter \(\text{MoA}_\text{attn}\) facilitates information stitch between feature maps from the \(i\)-th and \(j\)-th modalities, allowing cross-modal interaction:
\begin{equation}
x_{i,\text{adapt}_1}^k = \text{MoA}_{\text{attn}}(x_{i,\text{attn}}^k, x_{j,\text{attn}}^k),
\end{equation}
for \(i \neq j\). Each input is then processed independently through the Feature-Forward Networks (FFN) module. This step involves non-linear transformations that further refine the features, preparing them for subsequent stages in the model:
\begin{equation}
x_{i,\text{ffn}}^k = x_{i,\text{adapt}_1}^k + \text{DP}(\text{FFN}(\text{LN}_2(x_{i,\text{adapt}_{1}}^k))).
\end{equation}

Following the FFN, the modality adapter \(\text{MoA}_\text{ffn}\) is applied again to facilitate final cross-modal information stitch:
\begin{equation}
x_{i,\text{adapt}_2}^k = \text{MoA}_\text{ffn}(x_{i,\text{ffn}}^k, x_{j,\text{ffn}}^k),
\end{equation}
for \(i \neq j\). This second application of the modality adapter guarantees that the refined features from different modalities are effectively integrated into the model.

\begin{figure*}
\centering
\includegraphics[width=0.9\linewidth]{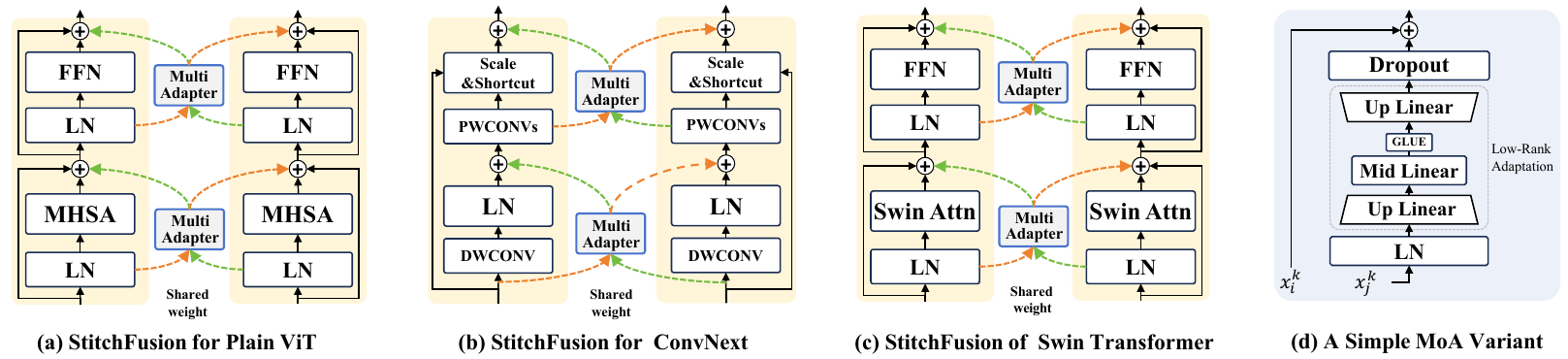} 
\vspace{-10pt}
\caption{StitchFusion Framework For Different Encoder Architectures. We use a simple low-rank adaptation module as our MoA.}
\label{fig:fig_3}
\vspace{-15pt}
\end{figure*}

The structure and sequence of operations are illustrated in \cref{fig:fig_3}(a), with the corresponding pseudocode provided in Algorithm \ref{StitchFusion_pseu}. Additional schematic representations for Swin Transformer and ConvNeXt models are also shown in \cref{fig:fig_3}(b-c); however, these are not elaborated here and can be further reviewed in the pseudocode included in the Appendix \textcolor[rgb]{0.4, 0, 0.4}{B}.

\begin{algorithm}
\small
\caption{StitchFusion Pseudocode}
\label{StitchFusion_pseu}
\begin{algorithmic}
\STATE \textbf{Input:} Input feature maps $x^k$ for the $k$-th encoder block
\STATE \textbf{Output:} Fused feature map $x^{k+1}$

\STATE // MHSA
\STATE \textbf{For each} $i$ modality:
\STATE \quad $x_{i,\text{attn}}^k \leftarrow x_i^k + \text{DropPath}(\text{MHSA}(\text{LN}_1(x_i^k)))$
\STATE \textbf{For each} $i$ modality:
\STATE \quad \textbf{For each} $j$ modality:
\STATE \quad\quad \textbf{if} $i \neq j$:
\STATE \quad\quad\quad $x_{i,\text{adapt}_1}^k \leftarrow \text{MoA}_{\text{attn}}(x_{i,\text{attn}}^k, x_{j,\text{attn}}^k)$

\STATE // FFN
\STATE \textbf{For each} $i$ modality:
\STATE \quad $x_{i,\text{ffn}}^k \leftarrow x_{i,\text{adapt}_1}^k + \text{Dropout}(\text{FFN}(\text{LN}_2(x_{i,\text{adapt}_1}^k)))$
\STATE \textbf{For each} $i$ modality:
\STATE \quad \textbf{For each} $j$ modality:
\STATE \quad\quad \textbf{if} $i \neq j$:
\STATE \quad\quad\quad $x_{i,\text{adapt}_2}^k \leftarrow \text{MoA}_{\text{ffn}}(x_{i,\text{ffn}}^k, x_{j,\text{ffn}}^k)$

\STATE \textbf{Equivalence}: $x^{k+1} \leftarrow x_{i,\text{adapt}_2}^k$

\STATE \textbf{Return:} Final fused feature map $x^{k+1}$
\end{algorithmic}
\end{algorithm}

\subsubsection{Multi-directional Modality Adapter For Information Fusion}

In the realm of multi-modal learning, effectively fusing information from diverse sources is paramount. We consider parameter efficiency in this chapter and introduce a novel component termed the Modality Adapter, denoted as \( \text{MoA}_\text{attn} \) and  \( \text{MoA}_\text{ffn} \) in last section, which serves as information stitcher. This adapter is designed simply based on a linear module augmented with non-linear transformations.

The Modality Adapter performs a sequence of operations on the input data, each tailored to refine and prepare the features for subsequent tasks. These operations include Downscaling, Processing, and Upscaling, which we detail below.

\subsubsection{Downscaling.}
For the $k$-th block, the initial stage involves a downscaling transformation aimed at reducing the spatial dimensions of the input feature \( x^k \in \mathbb{R}^{H_kW_k \times C_k} \) while retaining essential information. This is achieved through a linear transformation:
\begin{equation}
    x_{j,\text{down}}^k = \mathbf{W}_\text{down} \cdot x_j^k + \mathbf{b}_\text{down},
\end{equation}
Here, \( \mathbf{W}_\text{down} \in \mathbb{R}^{H_kW_k \times r} \) is the trainable weight matrix and \( \mathbf{b}_\text{down} \in \mathbb{R}^{r} \) is the corresponding bias vector, both of which are crucial for learning the appropriate feature compression. $r$ is the number of hidden dimension.

\subsubsection{Processing.}
The downscaled features are then subjected to further processing. To introduce non-linearity and enhance the model's representational power, we employ a GELU activation function. Additionally, to prevent overfitting and promote generalization, dropout regularization is applied:
\begin{equation}
    x_{j,\text{mid}}^k = \text{Dropout}(\text{GELU}(\mathbf{W}_\text{mid} \cdot x_{j,\text{down}}^k + \mathbf{b}_\text{mid})),
\end{equation}
In this equation, \( \mathbf{W}_\text{mid} \in \mathbb{R}^{r \times r}  \) and \( \mathbf{b}_\text{mid} \in \mathbb{R}^{r}\) are the weight matrix and bias vector, respectively, that operate on the downscaled features, shaping them for the final upscaling phase.

\subsubsection{Upscaling.}
The final operation is the upscaling, which is the inverse of the downscaling process. It restores the feature dimensions to their original size, allowing the model to make fine-grained predictions:
\begin{equation}
    x_{j,\text{up}}^k = \text{DropPath}(\mathbf{W}_\text{up} \cdot x_{j,\text{mid}}^k + \mathbf{b}_\text{up}),
\end{equation}
The weight matrix \( \mathbf{W}_\text{up} \in \mathbb{R}^{r \times H_kW_k} \) and bias vector \( \mathbf{b}_\text{up} \in \mathbb{R}^{H_kW_k}\) are responsible for this dimensional restoration, ensuring that the upscaled features are well-aligned with the input space.

\subsubsection{Fusion.}
For $i$-th modaltiy feature $x_i^k$ and the projected $j$-th modaltiy feature $x_{j,\text{up}}^k$, we perform a simple addition operation for feature fusion.
\begin{equation}
    x_{i,\text{adapt}}^k = x_i^k + x_{j,\text{up}}^k,
\end{equation}
The fused feature $x_{i,\text{adapt}}^k$ is used for afterward processing.

\subsubsection{Why a simple MLP.}
We only concentrate on exploring the role of the proposed new modality fusion perspective. Therefore, this article only employs a relatively simple and useful modality fusion device, namely, a simple low-rank adaptation module inspired by \cite{hu2022lora}. As for the novel Modality Adapter design, we do not take it into account for the time being and the related research will be carried out in the subsequent work. 

\subsection{Modality Adapter At Different Density Levels}
This section extends the concept to support multiple modalities and different levels of connection density, providing a detailed analysis of various configurations and their implications for model performance. We supply the Illustration in Appendix \textcolor[rgb]{0.4, 0, 0.4}{D}.

\subsubsection{Shared  Modality Adapter for All Modalities.}
In this configuration, all modalities share the same  Modality Adapter (named \textit{sMoA}). This means that the same set of weights and biases are used for the transformations between any pair of modalities. This approach promotes consistency and reduces the overall number of parameters, making the model more efficient and easier to train. The transformation for any modality pair $(i, j)$ uses the same weights and biases:
\begin{equation}
    \mathbf{y}_{i \rightarrow j} = \text{sMoA}(x_i^k,x_j^k),
\end{equation}
\begin{equation}
    \mathbf{y}_{j \rightarrow i} = \text{sMoA}(x_j^k,x_i^k).
\end{equation}

Using a shared set of weights, the sMOA ensures that the transformations are uniform across all modality pairs, which can be beneficial in scenarios where the modalities have similar feature distributions. 

\subsubsection{Independent Modality Adapter for Each Pair of Modalities.} 
In this configuration, each pair of modalities has its own bi-directional  Modality Adapters (named obMoA). For $M$ modalities, there are $C_{M}^2$  Modality Adapters in total, where each pair of modalities $(i, j)$ is assigned a unique set of weight matrices and biases.  Let $x_i^k$ and $x_j^k$ be the feature vectors for modalities $i$ and $j$ respectively, the transformation is defined as:
\begin{equation}
    \mathbf{y}_{i \rightarrow j} = \text{obMoA}_{i \leftrightarrow j}(x_i^k,x_j^k),
\end{equation}
\begin{equation}
    \mathbf{y}_{j \rightarrow i} = \text{obMoA}_{i \leftrightarrow j}(x_j^k,x_i^k).
\end{equation}
Each pair $(i, j)$ has a unique set of weight matrices and biases.
This configuration allows for more specialized transformations tailored to the specific characteristics of each modality pair. This is especially useful when the modalities have significantly different feature distributions, as it allows for more precise adaptations.

\subsubsection{Parallel  Modality Adapters for Each Pair of Modalities.} 
In this configuration, each pair of modalities shares two uni-directional  Modality Adapter Modules (named \textit{tuMoA}). This means that there are separate weight matrices and biases for the transformations in each direction between two modalities, allowing for asymmetric information stitch. The transformation is:
\begin{equation}
    \mathbf{y}_{i \rightarrow j} = \text{tuMoA}_{i \rightarrow j}(x_i^k,x_j^k),
\end{equation}
\begin{equation}
    \mathbf{y}_{j \rightarrow i} = \text{tuMoA}_{j \rightarrow i}(x_j^k,x_i^k).
\end{equation}
where each pair $(i, j)$ has a shared set of weight matrices and biases. This setup can capture directional dependencies and interactions more effectively, as it can learn distinct transformations for each direction.

\subsubsection{Equivalence of Configurations for Two Modalities.} 
When the number of modalities $m = 2$, the configurations of a sMoA and an obMoA are equivalent. In both cases, the transformation involves a single set of weights and biases:
Thus, for \(m=2\):
\begin{equation}
\text{obMoA}_{1 \leftrightarrow 2} = \text{sMoA}.
\end{equation}
In the remainder of the text, we will refer to the framework using \textit{sMoA} as \textit{sStitchFusion}, the framework using \textit{obMoA} as \textit{obStitchFusion}, and the framework using \textit{tuMoA} as \textit{tuStitchFusion}. Each of these frameworks offers distinct advantages depending on the number of modalities and the specific requirements of the application, providing flexible options for multimodal integration.

%% file: section/04_experiments.tex
\section{Experimental Result}
\label{experiments}

\input{table/01_all_sota_1}

\subsection{Experimental Details}
\quad
We test our model on 7 datasets. The FFMs in the paper are configured as the module from \cite{kaykobad2023multimodal}. We set the intermediate dimention for MoA to \( r = 8 \) . the learning rate is \( 1.2 \times 10^{-4} \) for the FMB dataset and \( 6 \times 10^{-5} \) for others. Furthermore, we use cross-entropy as our training loss, and a warm-up technology is implemented for the initial 10 epochs, followed by a learning rate decay factor of 0.01. The StitchFusion configured with obMoA serves as the default model for our experiment.

\subsection{DataSet}
\quad
We select MCubeS Dataset \cite{liang2022multimodal}, FMB Dataset \cite{liu2023multi}, MFNet Dataset \cite{ha2017mfnet}, DeLiVER Dataset \cite{zhang2023delivering}, PST900 Dataset \cite{shivakumar2020pst900}, NYUv2 \cite{Silberman:ECCV12}, SUN\cite{song2015sun} as our test benchmark, the introduction of dataset please refer to the Appendix \textcolor[rgb]{0.4, 0, 0.4}{C}. Moreover, we introduce a multimodality underwater segmentation dataset (MMUS) and test our StitchFusion on it, 

\subsection{Experimental Results on Datasets}
The results of our StitchFusion model, as shown in Tables \ref{tab:sota_results1} and \ref{tab:sota_results2}, demonstrate its effectiveness in multimodal semantic segmentation across diverse datasets, while also highlighting certain limitations.
On the NYUDv2 dataset, it attains high scores with various backbones. Notably, when employing the Swin-Large-22k backbone, its performance ranks second only to that of GeminiFusion, but without additional strategy. On the DeLiVER dataset, it outperforms all compared methods, achieving the highest mIoU of 70.3 with the Swin-Tiny-1k backbone. Similarly, on the Mcubes dataset, StitchFusion achieves the best performance with an mIoU of 55.9 using the Swin-Large-22k backbone. On the SUN dataset, StitchFusion variants with different backbones (MiT-B5, Swin-Tiny-1k, Swin-Large-22k) achieve accuracies of 53.4, 50.3, and 54.8 respectively, showing competitive performance. Additionally, for the FMB and PST900 datasets, StitchFusion significantly surpasses existing methods, with 68.7 mIoU on FMB and 89.4 mIoU on PST900, setting a new benchmark for these datasets. 
Nevertheless, the model also has certain limitations. On the NYUDv2 dataset, compared to GeminiFusion, it has a marginally lower score (59.6 vs 60.9) when using the same Swin-Large-22k backbone. This implies potential inadequacies in either detail feature extraction for RGB-D modality combination, therefore we make a multimodal underwater segmentation (MMUS) dataset to reexamine the RGB-D adaptation ability of StitchFusion (we illustrate the dataset and results in the Appendix \textcolor[rgb]{0.4, 0, 0.4}{A}). Besides, we append the Per-Class mIoU results for some datasets in the Appendix \textcolor[rgb]{0.4, 0, 0.4}{F}.

\input{table/02_all_sota_2}
\subsection{Ablation Experiment}

\subsubsection{Comparison of different modal combinations.}
\input{table/03_different_modality}
The results in Table \ref{tab:Different_modality_Mcubes} demonstrate that StitchFusion effectively leverages multimodal inputs to enhance segmentation performance. On the MCubeS dataset, incorporating additional modalities consistently improves accuracy, with the best performance of 53.92 achieved using RGB-A-D-N and FFMs with the MiT-B4 backbone. Similarly, on the DeLiVER dataset, the RGB-D-E-L combination yields the highest accuracy of 70.34 with Swin-Tiny-1k, showcasing the model's ability to fuse diverse information for complex scenes.
\subsubsection{Exploration of configuration density for different StitchFusion}
\input{table/04_Ablation_density_diff_stage} 
The results in Table \ref{tab:Ablation_diff_stage} evaluate the impact of deploying StitchFusion at different stages. Performance improves progressively as the model integrates multimodal information across later stages, with the highest accuracy observed when StitchFusion is applied at all stages.

\subsubsection{Exploration of Different Variant of StitchFusion}
\input{table/04_Ablation_different_variant_compatible}
The Table \ref{tab:Ablation_variants} show that ablation study on various StitchFusion variants. It demonstrates the adaptability of the proposed model to different configurations. As observed, the standard StitchFusion achieves robust performance across multiple datasets, indicating its foundational strength. Variants provide slight trade-offs in performance, revealing the influence of structural adjustments.

\subsubsection{StitchFusion is compatible with existing Feature Fusion Modules.}
The integration of FFMs with StitchFusion yields significant performance improvements, as reflected in the mIoU scores in Table \ref{tab:Ablation_variants}. The inclusion of FFMs enhances feature refinement and alignment, effectively complementing StitchFusion's architecture. 

\subsubsection{Exploration of Hidden Dimension for StitchFusion}
\input{table/05_Ablation_hiddendim}
The ablation study on hidden dimension $r$ reveals the impact of dimensionality on model performance and parameter efficiency. As shown in Table~\ref{hidden_dim}, increasing $r$ generally improves performance, but the gain saturates beyond a certain point. Specifically, hidden dimension values of 4 and 8 strike an optimal balance, achieving competitive mIoU scores of 49.46\% and 49.56\% on Mcubes-RGBN and 83.29\% and 83.82\% on PST900-RGBT, respectively, while maintaining a low parameter footprint of 0.47\% and 0.89\%. Higher hidden dimension values, such as 64 or 256, slightly increase performance but result in disproportionately large parameter growth. Therefore, $r$ = 4 and 8 are recommended as they provide a favorable trade-off between accuracy and computational cost, demonstrating the efficiency of StitchFusion in managing model complexity.

\subsection{Parameter Number Efficiency}
\subsubsection{Parameter Number Analysis.}
To highlight the efficiency of the StitchFusion module in terms of parameter number, we present the following formula for calculating the module's parameter number \(\textit{P}\):
\begin{equation}
\begin{aligned}
\textit{P}
    &= \sum_{i} \left( (2rd_i + r^2) \cdot 2 \cdot C_{m}^2 \cdot \sum_{j} (\text{$depth$}_{i,j})\right),
\end{aligned}
\end{equation}
where \( d_i \) denotes the feature dimensions of input and output for the \( i \)-th stage, \( r \) is a hyperparameter that adjusts StitchFusion's downsampling dimensions (set to 8 in this paper), \( m \) represents the number of modalities, and \(depth_{i,j}\) indicates the depth of the backbone network for the \( i \)-th stage, \(C_{m}^2\) represents the number of combinations when choosing 2 encoders to perform information-sharing out of m distinct modalities.

\subsubsection{Parameter Number Addition Analysis.}
The parameter efficiency of StitchFusion demonstrates its balance between performance (mIoU) and computational cost. As shown in Table~\ref{tab:params_DE_MC}, StitchFusion consistently outperforms state-of-the-art methods like CMNeXt and MMSFormer with fewer parameters. For instance, on the DeLiVER dataset, StitchFusion (MiT-B2) achieves 68.18\% mIoU with 26.50M parameters, surpassing CMNeXt's 66.30\% with 58.73M parameters. Similarly, on the Mcubes dataset, StitchFusion (MiT-B4) achieves 53.92\% mIoU with 68.02M parameters, outperforming MMSFormer. Some more additional results are shown in Appendix \textcolor[rgb]{0.4, 0, 0.4}{G}. 

\subsection{Computational Complexity Analysis.}
Table~\ref{tab:Analysis_GFLOPs} shows the analysis of computational cost in StitchFusion. Table (a) demonstrates the impact of different numbers of modalities and resolutions on computational complexity. Table (b) indicates that when the number of modalities is small, StitchFusion exhibits a smaller increase in the number of parameters and a minimal increase in GFLOPs compared to the baseline Segformer. However, as the number of modalities increases, StitchFusion shows a relatively more significant increase in GFLOPs. In all, StitchFusion effectively balances multimodal performance with low computational costs. More analysis is shown in Appendix \textcolor[rgb]{0.4, 0, 0.4}{E}.

\input{table/06_Analysis_param_miou}
\input{table/08_Analysis_GFLOPs}

\subsection{Semantic Segmentation Visulization}

\begin{figure}[ht]
\centering
\vspace{-10pt}
\includegraphics[width=0.85\linewidth]{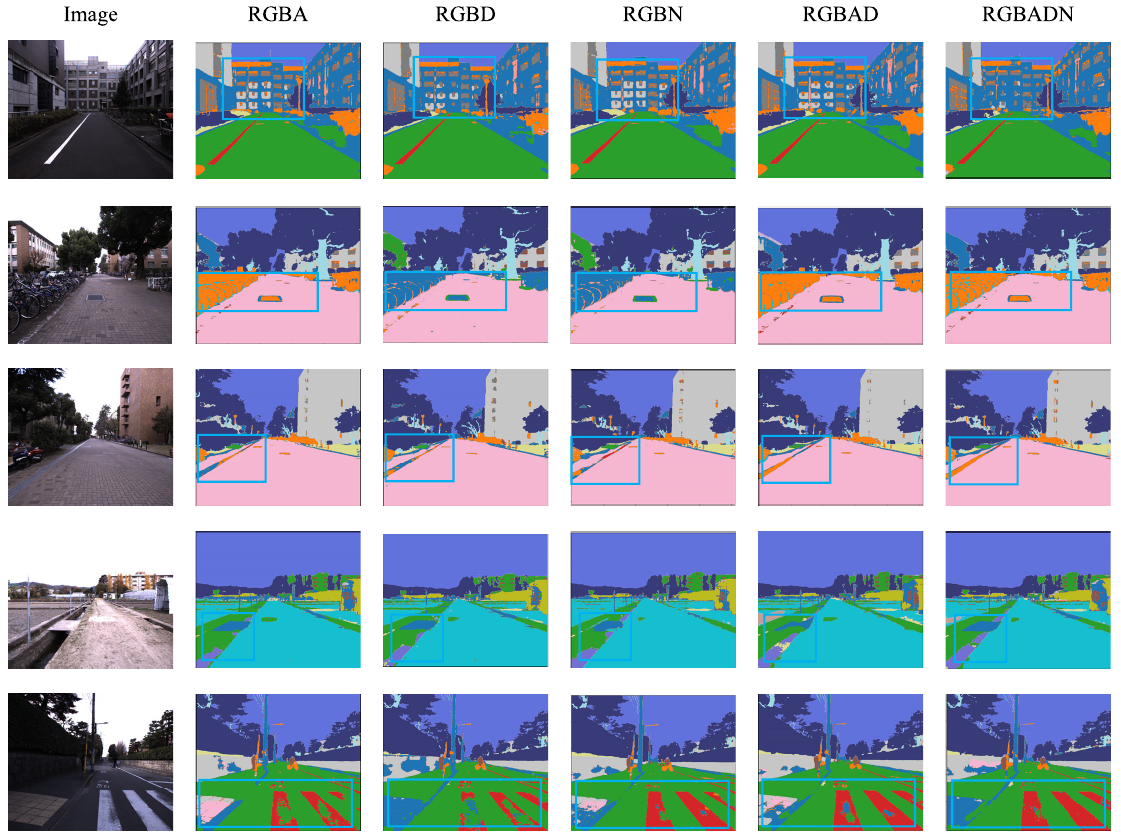}
\vspace{-10pt}
\caption{Visulization of StitchFusion (MiT-B4) On Mcubes Dataset.}
\label{fig:Visulization_mcubes_1}
\vspace{-15pt}
\end{figure}
\subsubsection{Segmentation Visualizations.} The segmentation visualizations in Fig.\ref{fig:Visulization_mcubes_1} highlight the effectiveness of StitchFusion on the Mcubes dataset and DeLiVR dataset. With more modalities (RGBA, RGBD, RGBN, RGBAD, RGBADN), the segmentation results become progressively more accurate and detailed. More visualization is shown in Appendix \textcolor[rgb]{0.4, 0, 0.4}{H}.

%% file: table/01_all_sota_1.tex
\begin{table*}[t]
\centering
\small
\caption{\small{Comparison results with state-of-the-art methods on the NYUDv2, DeLiVER and Mcubes datasets for the multimodal semantic segmentation task. MiT-B5 for SUN Dataset, MiT-B2 for DeLiVER Dataset and MiT-B4 for other Dataset. '-' in our table indicates that our model is unable to generate results due to an Out of Memory (OOM) error.}
}
\label{tab:sota_results1}
\vspace{-10pt}
\resizebox{0.80\linewidth}{!}
{
\begin{tabular}{l|l|l|c|c|c|c|c}
\toprule
Method & Backbone & Publication & Additional Strategies  & NYUDv2 & DeLiVER & Mcubes & SUN\\
\midrule
TokenFusion      & MiT-B$5$ (MiT-B$2$) & CVPR2022 \cite{wang2022multimodal} & \xmark     & 55.1          & 63.5       & -    & 53.0 \\
SMMCL  & SegNeXt-B & WACV2024 \cite{Dong2024SMMCL} & \xmark     & 55.8          & -          & -      & - \\
MultiMAE & ViT-Base & ECCV2022 \cite{bachmann2022multimae} & \cmark     & 56.0          & -          & -    & - \\
OMNIVORE & Swin-Large & CVPR2022 \cite{girdhar2022omnivore} & \cmark     & 56.8          & -          & -     & - \\
CMNeXt & MiT-B$4$ (MiT-B$2$) & CVPR2023 \cite{zhang2023delivering} & \xmark     & 56.9          & 66.3       & 51.5    \\
CMX & MiT-B$5$ & TITS2023 \cite{zhang2023cmx} & \xmark     & 56.9          & 62.7       & -   & 52.4 \\
DFormer & DFormer-L & ICLR2024 \cite{yin2023dformer} & \cmark     & 57.2          & -          & -      & 52.5 \\
PolyMaX  & ConvNeXt-L & CVPR2024 \cite{yang2024polymax} & \cmark     & 58.1          & -          & -      & - \\
SwinMTL & SwinV2-Base-MiM & IROS2024 \cite{taghavi2024swinmtl} & \cmark     & 58.1          & -          & -      &  - \\
EMSANet & EMSANet-R34-NBt1D & IJCNN2022 \cite{seichter2022efficient} & \cmark     & 59.0          & -          & -      & 48.5 \\
DPLNet & MiT-B5 & IROS2024 \cite{dong2023efficient} & \cmark     & 59.3          & -          & -      & - \\
GeminiFusion & MiT-B$5$ (MiT-B$2$) & ICML2024 \cite{jia2024geminifusion} & \xmark     & 57.7          & 66.9       & -  & 53.3 \\
GeminiFusion & Swin-Large-22k & ICML2024 \cite{jia2024geminifusion} & \xmark     & \textbf{60.2}          & -          & -     & \underline{54.6}  \\
MCubeSNet & MiT-B4 & CVPR2022 \cite{liang2022multimodal} & \xmark     & -             & -         & 42.9  & - \\
ShareCMP & MiT-B2 & arXiv2022 \cite{liu2023sharecmp} & \xmark     & -             & -         & 50.3  & -  \\
DeepLabV3+  & ResNet-$101$ & ECCV2018 \cite{chen2018encoder} & \xmark     & -             & -         & 38.1  & -  \\
MMSFormer & MiT-B4 & IOJSP2023 \cite{kaykobad2023multimodal} & \xmark     & -             & -         & 53.1 & -  \\
\midrule
StitchFusion & MiT-B2 (-B4,-B5) & ACMMM2025 & \xmark         & 57.8          & \underline{68.2}  & \underline{53.9}   & 53.4  \\
StitchFusion & Swin-Tiny-1k (-22k) & ACMMM2025 & \xmark          & 53.8          & \textbf{70.3}     & 52.3    & 50.3 \\
StitchFusion & Swin-Large-22k & ACMMM2025 & \xmark        & \underline{59.6}       & -        & \textbf{55.9}      & \textbf{54.8} \\
\bottomrule
\end{tabular}
}
\vspace{-10pt}
\end{table*}

%% file: table/02_all_sota_2.tex
\begin{table*}[t]
\centering
\small
\caption{\small{Comparison results with state-of-the-art methods on the FMB, PST900 and MFNet datasets.}}
\label{tab:sota_results2}
\vspace{-10pt}
\resizebox{0.75\linewidth}{!}
{
\begin{tabular}{l|l|l|c|c|c|c}
\toprule
Method  & Backbone & Publication & Additional Strategies  & FMB & PST900 & MFNet \\
\midrule   
MFNet            & -          & IROS2017 \cite{ha2017mfnet} & \xmark            & -    & 57.0      & 39.7      \\
RTFNet             & -        & IRAL2019 \cite{sun2019rtfnet} & \xmark            & -     & 57.6    & 53.2       \\
FuseSeg          & - & TASE2019 \cite{sun2020fuseseg} & \xmark            & -     & -  & -     \\
EGFNet               & -           & AAAI2022 \cite{zhou2022edge} & \xmark            & -     & 78.5    & 54.8        \\
ABMDRNet   & -            & CVPR2021 \cite{zhang2021abmdrnet} & \xmark            & -     & 67.3     & 54.8     \\
ECGFNet         & -          & TITS2023 \cite{zhou2023embedded} & \xmark            & -     & -     & 55.3       \\
FEANet           & - & IROS2021 \cite{deng2021feanet} & \xmark            & 46.8     & 85.5  & 55.3     \\
ABMDRNet+ & -           & TNNLS2023 \cite{zhao2023mitigating} & \xmark            & -     & -  & 56.8     \\
CAINet             & mobilenet-v2           & TMM2023 \cite{lv2023cainet} & \xmark            & -     & 54.7     & 58.6       \\
EAEFNet          & -         & IRAL2023 \cite{liang2023explicit} & \cmark            & -     & 85.4     & 58.9       \\
CMX                & MiT-B4 (MiT-B3)     & TITS2023 \cite{zhang2023cmx} & \xmark            & -     & -     & 59.7       \\
HAPNet             & ConvNext-L              & arXiv2024 \cite{li2024hapnet} & \cmark            & -          & \underline{89.0}     & \textbf{61.2}       \\
U2Fusion      & VGG16       & TPAMI2020 \cite{xu2020u2fusion} & \xmark            & 47.9          & -  & - \\ 
TarDAL             & -        & CVPR2022 \cite{liu2022tardal} & \xmark            & 48.1          & - & - \\ 
SegMiF            & MiT-B4        & ICCV2023 \cite{liu2023segmif} & \xmark            & 54.8          & - & - \\
MMSFormer   & MiT-B4 (MiT-B3)        & IOJSP2023 \cite{kaykobad2023multimodal} & \xmark            & 61.7          & - & - \\ 
\midrule 
StitchFusion         & MiT-B4 (MiT-B3) & ACMMM2025 & \xmark & \underline{64.3}         & 84.7     & 57.8  \\
StitchFusion         & Swin-Tiny-1k (-22k)   & ACMMM2025 & \xmark & 63.1          & 83.4     & 55.9  \\
StitchFusion         & Swin-Large-22k   & ACMMM2025 & \xmark & \textbf{68.7}        & \textbf{89.4}     & \underline{60.7}  \\
\bottomrule
\end{tabular}
}
\end{table*}

%% file: table/03_different_modality.tex
\begin{table}[htbp]
\centering
\caption{Comparison on Different Modalities Combinations of StitchFusion On MCubeS and DeLiVER dataset.}
\label{tab:Different_modality_Mcubes}
\vspace{-10pt}
\resizebox{\linewidth}{!}{
\begin{tabular}{llcccccc}
\shline
\textbf{Methods} & \textbf{Backbone} & \textbf{RGB-N} & \textbf{RGB-D} & \textbf{RGB-A} & \textbf{RGB-A-D} & \textbf{RGB-A-D-N} &  \\ 
\shline
StitchFusion        & MiT-B4        & 51.67 & 51.25 & 52.08 & \textbf{52.52} & \underline{51.74} & -\\
StitchFusion+FFMs   & MiT-B4        & 53.21 & 52.72 & 52.68 & \underline{53.26} & \textbf{53.92} & -\\
StitchFusion        & Swin-Large-22k& 54.59 & 55.69 & 56.07 & \textbf{56.94} & \underline{55.96} & -\\
\shline
\textbf{Methods} & \textbf{Backbone} & \textbf{RGB-D} & \textbf{RGB-E} & \textbf{RGB-L} & \textbf{RGB-D-L} & \textbf{RGB-D-E} & \textbf{RGB-D-E-L} \\ 
\shline
StitchFusion        & MiT-B2        & 65.75 & 57.44 & 58.03 & 66.03 & \underline{66.65} & \textbf{68.18} \\
StitchFusion        & Swin-Tiny-1k  & 68.75 & 57.80 & 58.67 & 69.01 & \underline{69.75} & \textbf{70.34} \\
\shline
\end{tabular}}
\vspace{-5pt}
\end{table}

%% file: table/04_Ablation_density_diff_stage.tex
\begin{table}[htbp]
    \centering
    \caption{Comparison of different StitchFusion(Swin-Tiny-1K.) position settings. stage-X means that StitchFusion is set at the X-th stage.}
    \label{tab:Ablation_diff_stage}
    \vspace{-10pt}
    \resizebox{\linewidth}{!}{
    \begin{tabular}{lccccccc}
        \shline
        Dataset-Modals    & stage0    & stage1    & stage2    & stage3    & stage0-1  & stage2-3  & stage-All\\
        \shline
        MCUBES-RGBN       & 47.97     & 48.04     & 49.27     & 49.24     & 48.46     & \underline{49.56} & \textbf{49.72}  \\
        PST900-RGBT       & 82.09     & 82.02     & 83.23     & 83.31     & 83.13     & \underline{83.32} & \textbf{83.43}  \\
        FMB-RGBT          & 61.47     & 61.09     & 61.94     & 62.37     & 61.50     & \underline{62.81} & \textbf{63.12}  \\
        \shline
    \end{tabular}
    }
\end{table}

%% file: table/04_Ablation_different_variant_compatible.tex
\begin{table}[htbp]
\centering
\small
\caption{\small{Comparison of StitchFusion's Different Levels of Dense Connectivity on Mcubes and FMB Datasets. \dag denotes performing a pixel-wise summation of modalities other than RGB to create a new modality input. * denotes the  Modality Adapter is used in the latter two stages. SF stands for StitchFusion.}}
\label{tab:Ablation_variants}
\resizebox{\linewidth}{!}{
\begin{tabular}{llc|llc}
\shline
\multicolumn{3}{c|}{\textbf{Mcubes Dataset}} & \multicolumn{3}{c}{\textbf{FMB Dataset}} \\ 
\shline
\textbf{Methods} & \textbf{Backbone} & \textbf{mIoU (\%)} & \textbf{Methods} & \textbf{Backbone} & \textbf{mIoU (\%)} \\ 
\shline
sStitchFusion       & MiT-B4    & 51.50             & StitchFusion           & MiT-B3   & \textbf{63.30} \\
StitchFusion\dag    & MiT-B4    & 51.22             & StitchFusion*         & MiT-B3    & 62.28 \\
StitchFusion        & MiT-B4    & \textbf{51.70}    & tuStitchFusion        & MiT-B3    & 62.82 \\
tuStitchFusion      & MiT-B4    & 51.12             & tuStitchFusion*       & MiT-B3    & 62.64 \\
\shline
sSF+FFMs            & MiT-B4    & 51.94             & SF+FFMs           & MiT-B3    & 64.32 \\
SF\dag+FFMs         & MiT-B4    & 52.14             & SF*+FFMs          & MiT-B3    & 64.51 \\
SF+FFMs             & MiT-B4    & \textbf{53.92}    & tuSF+FFMs         & MiT-B3    & 63.25 \\
tuSF+FFMs           & MiT-B4    & 51.89             & tuSF*+FFMs        & MiT-B3    & \textbf{64.85} \\
\shline
\multicolumn{3}{c|}{\textbf{MFNet Dataset}} & \multicolumn{3}{c}{\textbf{PST900 Dataset}} \\ 
\shline
\textbf{Methods} & \textbf{Backbone} & \textbf{mIoU (\%)} & \textbf{Methods} & \textbf{Backbone} & \textbf{mIoU (\%)} \\ 
\shline
StitchFusion    & MiT-B4    &57.80                   & StitchFusion    & MiT-B4    &84.70 \\
StitchFusion*    & MiT-B4    &57.76                  & StitchFusion*    & MiT-B4    &83.41 \\
\shline
StitchFusion+FFMs    & MiT-B4    &57.91                   & StitchFusion+FFMs    & MiT-B4    &85.35 \\
StitchFusion*+FFMs    & MiT-B4    &58.13                  & StitchFusion*+FFMs    & MiT-B4    &85.31\\
\shline
\end{tabular}
}
\end{table}

%% file: table/05_Ablation_hiddendim.tex
\begin{table}[htbp]
    \centering
    \small
    \vspace{-10pt}
    \caption{Comparision of Different hidden dimension of StitchFusion. Since the differences in experimental results are not significant, we recommend using $r$ = 4/8 (in \textbf{bold}), as this balances high precision and the increase in parameters.} 
    \label{hidden_dim}
    \vspace{-10pt}
    \resizebox{\linewidth}{!}{
    \begin{tabular}{lcccccc}
        \shline
        HiddenDim & 1 & 4 & 8 & 16 & 64 & 256 \\
        \shline
        Mcubes-RGBN(Swin-Tiny-1k) & 48.29 & \textbf{49.46} & \textbf{49.56} & 49.81 & 49.12 & 49.34 \\
        PST900-RGBT(Swin-Tiny-1k) & 82.12 & \textbf{83.29} & \textbf{83.82} & 83.52 & 83.61 & 83.80 \\
        \hdashline
        Params. Percentage & 0.16\%	& \textbf{0.47\%} 	& \textbf{0.89\%}  &1.73\%	& 6.73\%	&25.42\% \\
        \shline
    \end{tabular}
    }
    \vspace{-10pt}
\end{table}

%% file: table/06_Analysis_param_miou.tex
\begin{table}[htbp]
\centering
\small
\caption{Comparison of Parameters Efficiency With SOTA Methods On DeLiVER and Mcubes Dataset.}
\label{tab:params_DE_MC}
\vspace{-10pt}
\resizebox{\linewidth}{!}{
\begin{tabular}{llccc}
    \shline
    \textbf{DELIVER Dataset} & \textbf{Method} & \textbf{RGB-D} & \textbf{RGB-DE} & \textbf{RGB-DEL} \\
    \shline
    \multirow{2}{*}{Params (M)}  
    & CMNeXt(MiT-B2)  & 58.69  & 58.72  & 58.73 \\
    & StitchFusion(MiT-B2) & 25.93  & 26.22  & 26.50 \\
    \hdashline
    \multirow{2}{*}{mIoU. (\%)}
    & CMNeXt(MiT-B2)& 63.58  & 64.44  & 66.30 \\
    & StitchFusion(MiT-B2) & 65.75  & 66.03  & 68.18 \\
    \shline
    \textbf{Mcubes Dataset} & \textbf{Method} & \textbf{RGB-A} & \textbf{RGB-AD} & \textbf{RGB-ADN} \\
    \shline
    \multirow{2}{*}{Params (M)}  
    & MMSFormer(MiT-B4)  & 64.88  & 65.27  & 65.65 \\
    & StitchFusion(MiT-B4) & 65.28  & 66.45  & 68.02 \\
    \hdashline
    \multirow{2}{*}{mIoU. (\%)}
    & MMSFormer(MiT-B4) & 51.30  & 52.03  & 53.11 \\
    & StitchFusion(MiT-B4) & 52.68  & 53.26  & 53.92 \\
    \shline
\end{tabular}
}
\end{table}

%% file: table/08_Analysis_GFLOPs.tex
\begin{table}[htbp]
\centering
\vspace{-10pt}
\caption{Comparisons of Different Models in GFLOPs and Parameters} 
\label{tab:Analysis_GFLOPs}
\vspace{-10pt}
\resizebox{0.92\linewidth}{!}{
\begin{tabular}{c|lcccc}
\shline 
& \textbf{Modals Number} & \textbf{Resolution} & \textbf{Backbone} & \textbf{\#Params(M)} & \textbf{GFLOPs(G)} \\  
\shline 
\multirow{4}{*}{(a)} 
&2 & 512×512 & MiT-B4 & - & 127.00 \\
&3 & 512×512 & MiT-B4 & - & 184.00 \\
&4 & 512×512 & MiT-B4 & - & 242.00 \\
&2 & 800×600 & MiT-B3 & - & 93.49 \\
\shline 
& \textbf{Model-Modals} & \textbf{Resolution} & \textbf{Backbone} & \textbf{\#Params(M)} & \textbf{GFLOPs(G)} \\  
\shline 
\multirow{6}{*}{(b)}
&Segformer-RGBL        & 1024×1024     & MiT-B2    & 25.79 & 59.27 \\
&CMNeXt-RGBL           & 1024×1024     & MiT-B2    & 58.69 & 62.94\\
&StitchFusion-RGBL     & 1024×1024     & MiT-B2    & 25.93 & 59.93 \\
&Segformer-RGB-DEL     & 1024×1024     & MiT-B2    & 25.79 & 99.94 \\
&CMNeXt-RGB-DEL        & 1024×1024     & MiT-B2    & 58.73 & 65.42 \\
&StitchFusion-RGB-DEL  & 1024×1024     & MiT-B2    & 26.65 & 104.00 \\
\shline
\end{tabular}
}
\end{table}

%% file: section/05_conclusion.tex
\section{Conclusion}
\label{conclusion}
This paper presents StitchFusion, a novel framework for enhancing feature fusion and alignment in semantic segmentation. By progressively selecting and adapting relevant features. Experimental results demonstrate its superiority over existing methods. However, the current Modality Adapter, based on a simple low-rank adaption architecture, may lack efficiency for fine-grained feature processing designed for segmentation dataset. \textbf{Future work} will focus on designing a more effective and fine-grained Modality Adapter.
The StitchFusion's \textit{GFLOPs performance is not optimal}, because the StitchFusion process all modality data through the encoder. However, models using only RGB modality in pre-trained encoders \cite{zhang2023delivering, he2024prompting} achieve lower GFLOPs when the modality number increase. Future work will explore strategies to further reduce GFLOPs from this perspective.

%% file: section/06_appendix.tex
\newpage
\appendix
\newpage
\section{Multimodaltiy UnderWater Segmentation Dataset (MMUS) and Results}
\label{sec:MMUS}
Building on the traditional underwater semantic segmentation dataset, the SUIM dataset \cite{islam2020suim}, we make a simple \underline{M}ultiModality \underline{U}nderwater \underline{S}egmentation (MMUS) dataset. We employed Depth Anything-v2-Large \cite{depth_anything_v2} \cite{depth_anything_v1} as the depth map generation model to generate corresponding RGB-D data pairs.
We removed the mismatched data pairs from the training set and obtained 1,488 pairs of training set RGB-D data and 110 pairs of RGB-D test set training data.
The categories of these data are shown in the Table \ref{tab:category_MMUS}. We only used the test results of MiT-B3 and MiT-B4 to verify the effectiveness of our model for RGB-D modality fusion. 

\begin{table*}[h]
\centering
\caption{The Categories and Corresponding Colors for MMUS Dataset}
\label{tab:category_MMUS}
\resizebox{\linewidth}{!}{%
\begin{tabular}{l|l|l|l}
\shline
Class & Abbreviation & Color & Description\\
\shline
1 Background waterbody & BW & \# Black & Refers to the water area in the background.\\
2 Fish and vertebrates & FV & \# Red & Includes various fish species and other aquatic vertebrates.\\
3 Plants/sea-grass & PG & \# Green & Encompasses aquatic plants and seagrasses.\\
4 Robots/instruments & RI & \# Dark Blue & Various robots and instruments used for underwater operations, monitoring, or research.\\
5 Wrecks/ruins & WR & \# Yellow & Denotes sunken ships and various ruins left by human activities underwater.\\
6 Reefs and invertebrates & RE & \# Purple & Contains coral reefs and invertebrate aquatic animals.\\
7 Human divers & HD & \# Cyan & Refers to human divers who carry out activities underwater.\\
8 Sand/sea-floor (\& rocks) & SF & \# Light Blue & Signifies the sand on the seabed, the sea floor, and rocks that make up the seabed topography.\\
\shline
\end{tabular}
}
\end{table*}

Table \ref{tab:perclass_mmus_iou} shows the per-class mIoU comparison of the StitchFusion model with MiT-B3 and MiT-B4 backbones on the MMUS dataset for RGB-Depth modalities. For the MiT-B3 backbone, the mIoU increases from 66.33\% with RGB modality to 72.68\% with RGB-Depth modality, with significant improvements in multiple classes such as FV and RI. Similarly, for the MiT-B4 backbone, the mIoU rises from 72.32\% (RGB) to 74.66\% (RGB-Depth), with classes like WR and HD showing notable score increases. These results clearly demonstrate that the multimodal fusion of the StitchFusion model, by incorporating depth information, effectively enhances the model's performance. The additional geometric and spatial cues provided by the depth modality enable the model to capture more comprehensive information, leading to more accurate class-specific predictions and a higher overall mIoU.

\begin{table*}[ht]
\centering
\caption{Per-class mIoU (\%) comparison on MMUS dataset for RGB-Depth modalities of StitchFusion MiT-B3 and MiT-B4.
Bold represents the first, and underline represents the second.}
\label{tab:perclass_mmus_iou}
\resizebox{0.90\linewidth}{!}{
\begin{tabular}{lllccccccccc}
\shline
\textbf{Methods} & \textbf{Backbones} & \textbf{Modalities} & \textbf{BW} & \textbf{FV} & \textbf{PG} & \textbf{RI} & \textbf{WR} & \textbf{RE} & \textbf{HD} & \textbf{SF} & \textbf{mIoU(\%)} \\ \shline
StitchFusion  & MiT-B3 &  RGB      &  84.04 & 64.30 & 44.91 & 65.23 & 80.44 & 64.62 & 65.07 & 62.01 & 66.33 \\
StitchFusion  & MiT-B3 &  RGB-D      & 90.00 & 75.50& 29.37 & 82.64 & 85.46 & 71.63 & 75.72& 71.16 & \textbf{72.68} \\
\shline
StitchFusion  & MiT-B4 &  RGB      &  88.16 & 66.79 & 50.05 & 77.86 & 81.18 & 71.95 & 75.77 & 66.79 & 72.32 \\
StitchFusion  & MiT-B4 &  RGB-D      &  90.03 & 71.10 & 43.27 & 82.45& 87.03 & 73.12& 80.30 & 70.01 & \textbf{74.66}\\
\shline
\end{tabular}
}
\end{table*}

\begin{figure*}
\centering
\includegraphics[width=0.95\linewidth]{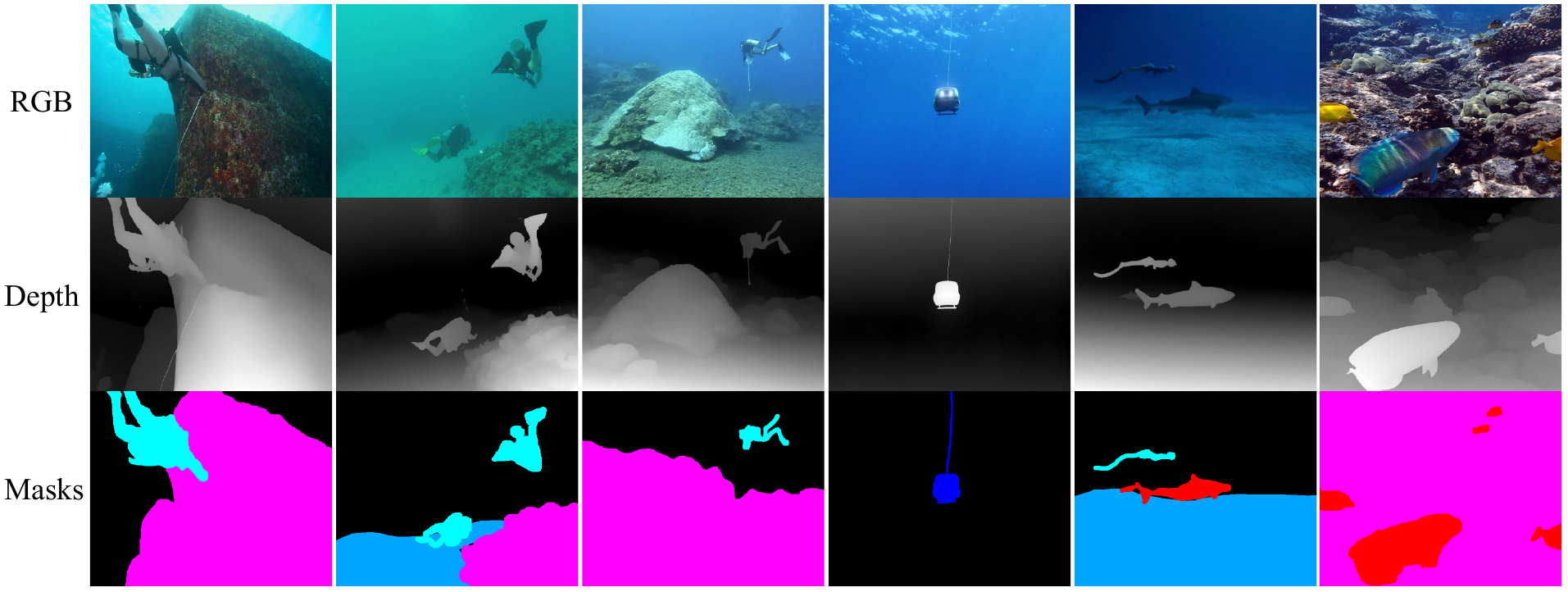} 
\vspace{-10pt}
\caption{Illustration of Different Modalities of MMUS.}
\label{fig:MMUS}
\end{figure*}

\section{Pseudocode Appendix}
\label{sec:pseudocode_app}
Since the specific workflow for Swin Transformer and ConvNeXt is not presented in the main text, we provide the pseudocode for the StitchFusion workflow with different encoders in the appendix.

We insert our $MOA$ in the way similar to the plain ViT method. The specific pseudocode is as follows.

\subsection{Pseudocode For Swin Transformer}

\begin{algorithm}
\caption{StitchFusion Pseudocode (Swin)}
\label{pseu_swin}
\resizebox{1.8\linewidth}{!}{%
\begin{minipage}{\textwidth}
\begin{algorithmic}
\STATE \textbf{Input:} Input feature maps $x^k$ for the $k$-th encoder block
\STATE \textbf{Output:} Fused feature map $x^{k+1}$
\STATE // MHSA
\STATE \textbf{For each} $i$ modality:
\STATE \quad $x_{i,\text{attn}}^k \leftarrow x_i^k + \text{DropPath}(\text{SwinAttn}(\text{LN}_1(x_i^k)))$
\STATE \textbf{For each} $i$ modality:
\STATE \quad \textbf{For each} $j$ modality:
\STATE \quad\quad \textbf{if} $i \neq j$:
\STATE \quad\quad\quad $x_{i,\text{adapt}_1}^k \leftarrow \text{MoA}_{\text{attn}}(x_{i,\text{attn}}^k, x_{j,\text{attn}}^k)$
\STATE // FFN
\STATE \textbf{For each} $i$ modality:
\STATE \quad $x_{i,\text{ffn}}^k \leftarrow x_{i,\text{adapt}_1}^k + \text{Dropout}(\text{FFN}(\text{LN}_2(x_{i,\text{adapt}_1}^k)))$
\STATE \textbf{For each} $i$ modality:
\STATE \quad \textbf{For each} $j$ modality:
\STATE \quad\quad \textbf{if} $i \neq j$:
\STATE \quad\quad\quad $x_{i,\text{adapt}_2}^k \leftarrow \text{MoA}_{\text{ffn}}(x_{i,\text{ffn}}^k, x_{j,\text{ffn}}^k)$
\STATE \textbf{Equivalence}: $x^{k+1} \leftarrow x_{i,\text{adapt}_2}^k$
\STATE \textbf{Return:} Final fused feature map $x^{k+1}$
\end{algorithmic}
\end{minipage}%
}
\end{algorithm}

\subsection{Pseudocode For ConvNext}

\begin{algorithm}
\small
\caption{StitchFusion Pseudocode (ConvNext)}
\label{pseu_conv}
\resizebox{1.85\linewidth}{!}{%
\begin{minipage}{\textwidth}
\begin{algorithmic}
\STATE \textbf{Input:} Input feature maps $x^k$ for the $k$-th encoder block
\STATE \textbf{Output:} Fused feature map $x^{k+1}$
\STATE // shortcut
\STATE \quad $x_{shortcut}^k \leftarrow x^k$

\STATE // dwconv
\STATE \textbf{For each} $i$ modality:
\STATE \quad $x_{i,\text{dw}}^k \leftarrow \text{LN}(\text{dwconv}(x_i^k))$
\STATE \textbf{For each} $i$ modality:
\STATE \quad \textbf{For each} $j$ modality:
\STATE \quad\quad \textbf{if} $i \neq j$:
\STATE \quad\quad\quad $x_{i,\text{adapt}_1}^k \leftarrow \text{MoA}_{\text{dw}}(x_{i,\text{dw}}^k, x_{j,\text{dw}}^k)$

\STATE // pwconv
\STATE \textbf{For each} $i$ modality:
\STATE \quad $x_{i,\text{pw}}^k \leftarrow x_{i,shortcut}^k + \text{DP}(\gamma \cdot \text{pwconv}_2(\text{GLUE}(\text{pwconv}_1(x_{i,\text{adapt}_1}^k)))) $
\STATE \textbf{For each} $i$ modality:
\STATE \quad \textbf{For each} $j$ modality:
\STATE \quad\quad \textbf{if} $i \neq j$:
\STATE \quad\quad\quad $x_{i,\text{adapt}_2}^k \leftarrow \text{MoA}_{\text{pw}}(x_{i,\text{pw}}^k, x_{j,\text{pw}}^k)$
\STATE \textbf{Equivalence}: $x^{k+1} \leftarrow x_{i,\text{adapt}_2}^k$
\STATE \textbf{Return:} Final fused feature map $x^{k+1}$
\end{algorithmic}
\end{minipage}%
}
\end{algorithm}

\section{Dataset}
\label{sec:appdataset}
\subsection{MCubeS Dataset.}
The MCubeS dataset includes RGB, Near-Infrared (NIR), Degree of Linear Polarization (DoLP), and Angle of Linear Polarization (AoLP) image pairs for semantic material segmentation across 20 categories. It consists of 302/96/102 image pairs for training/validation/testing, all sized at 1024×1024.

\subsection{FMB Dataset.}
The FMB dataset is designed for image fusion and segmentation, containing 1,500 infrared and visible image pairs annotated with 15 pixel-level categories. The training set has 1,220 pairs, and the test set has 280 pairs. 

\subsection{MFNet Dataset.}
The MFNet dataset stands out for its focus on thermal imagery in conjunction with RGB data, presenting a unique dataset of 1,569 image pairs. With a resolution of 640×480 pixels. The inclusion of 820 daytime and 749 nighttime image pairs provides a diverse set of conditions for training models to handle different lighting scenarios effectively. The MFNet dataset is segmented into 8 distinct classes, offering a rich ground for exploring the nuances of multi-modal segmentation in thermal contexts.

\subsection{DeLiVER Dataset.}
This dataset contains an impressive 47,310 frames with a subset of 7,885 annotated front-view samples. With images sized at 1024×1024 pixels, DeLiVER offers a high-resolution platform for developing advanced segmentation models that can leverage the depth of information provided by its multi-sensor approach. This dataset is particularly valuable for applications in autonomous driving, robotics, and any field requiring a comprehensive understanding of the environment from multiple perspectives.

\subsection{PST900 Dataset.} 
The PST900 dataset contains 894 synchronized RGB-Thermal image pairs with per-pixel ground truth annotations for five classes, divided into training and test sets.

\subsubsection{NYUv2 Dataset.} 
The NYUv2 RGB-depth dataset consists of a large collection of images captured in indoor environments, offering both RGB and depth data for each frame. The dataset includes 1,449 densely annotated scenes, with a total of 1449 RGB and depth pairs, providing ground truth labels for semantic segmentation.

\subsection{SUN Dataset.} 
The SUN RGB-D dataset, publicly released by the Vision\&Robotics Group at Princeton University, is designed for scene understanding tasks. It consists of 10335 real-world indoor images, with 5285 for training and 5050 for testing. The input resolution is 480x480, which is compatible with DFormer. The dataset encompasses 37 classes.

\section{Illustration of Different Density Modality Adapter}
\label{sec:fig_density_app}
We present the schematic diagrams of Modality Adapters with different densities in the appendix Fig.\ref{fig:fig_4}. Lines of different colors represent information of different modalities.

\section{Supplementary Computional Complexity Analysis}
\label{sec:CCA}
As for complexity. Our model reduces complexity from $\mathcal{O}(M^2\cdot N^2\cdot c)$ of cross-attention mechanism to $\mathcal{O}(M^2\cdot N\cdot c^2)$, improving efficiency. Here, $M$ is the number of modalities, $N$ is the number of tokens, and $c$ is the channels.

\begin{figure*}
\centering
\includegraphics[width=\linewidth]{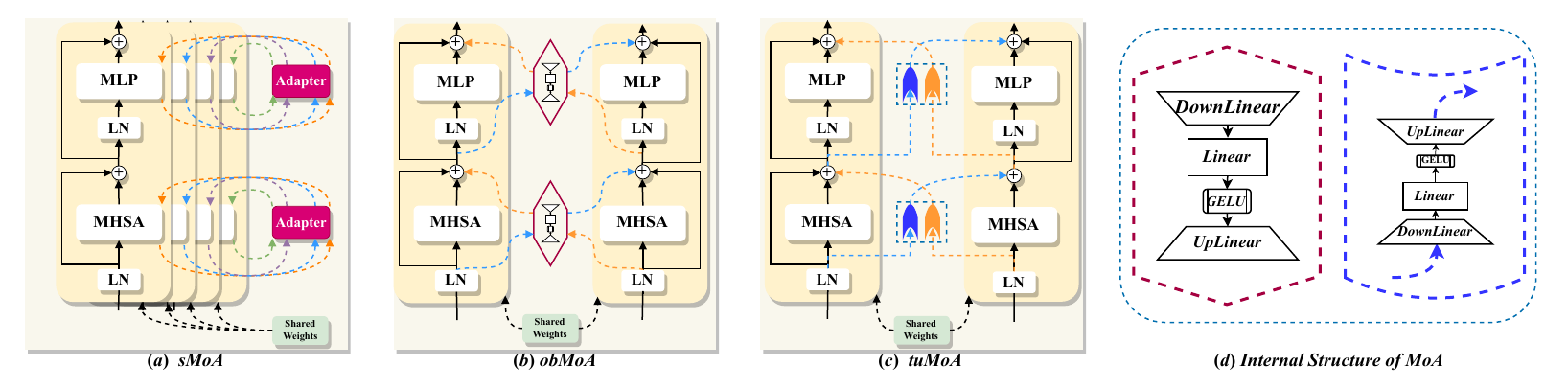}
\vspace{-22pt}
\caption{Modality Adapter Module At Different Density Levels. (a) Shared  Modality Adapter for All Modalities (\textit{sMoA}). (b) Independent  Modality Adapter for Each Pair of Modalities (\textit{obMoA}). (c) Parallel Modality Adapters for Each Pair of Modalities (\textit{tuMoA}). (d) A feasible internal structure of the Modality Adapter.}
\label{fig:fig_4}
\end{figure*}

\section{Per-class Comparision with State-of-the-art Models}
\label{sec:per_class_app}
\begin{figure*}[ht]
\centering
\includegraphics[width=0.95\linewidth]{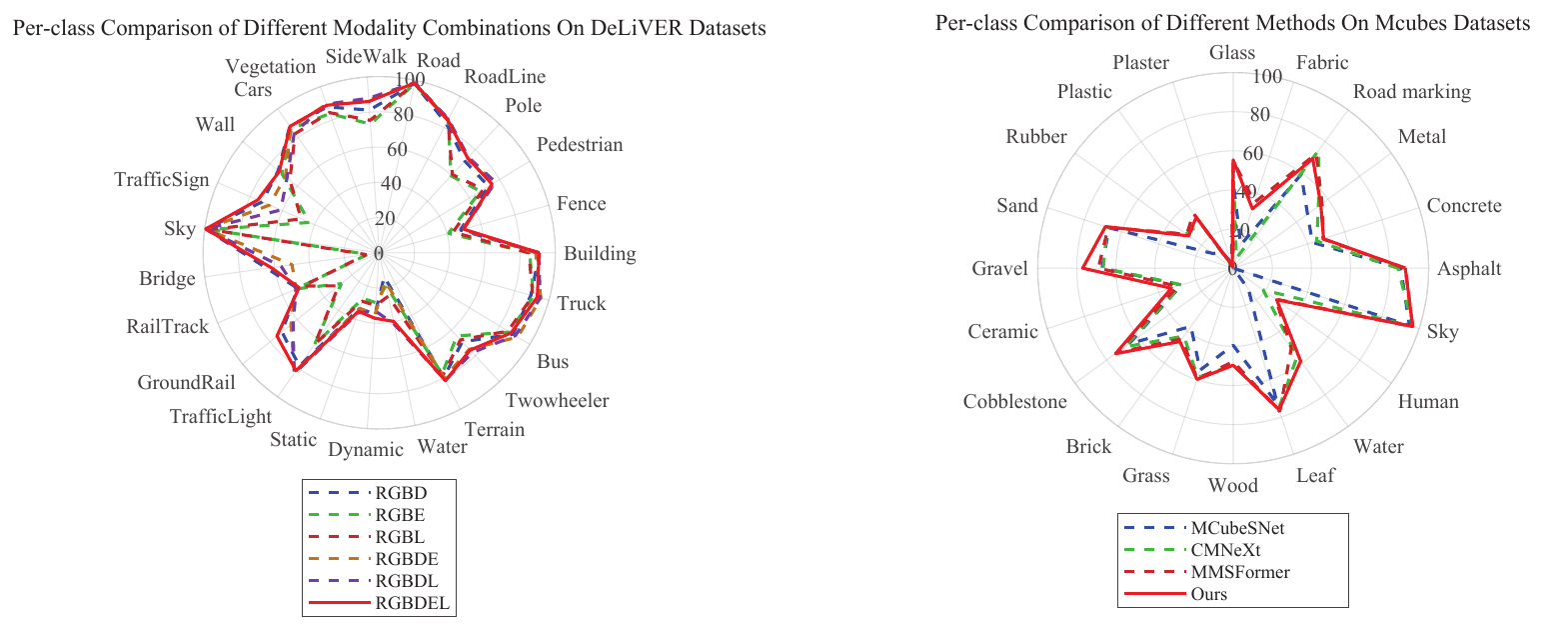}
\caption{Per-class IoU (\%) comparison on DeLiVER dataset (StitchFusion-MiT-B2) and Mcubes Dataset (StitchFusion-MiT-B4). Bold represents the first, and underline represents the second.}
\label{fig:perclass_lidar_fig}
\end{figure*}

In our analysis, we present a per-class comparative study of segmentation performance across datasets, focusing on the DeLiVER dataset and various modality combinations shown on the left side of Fig. \ref{fig:perclass_lidar_fig}. Additionally, we compare the per-class mIOU performance on the Mcubes (Table. \ref{fig:perclass_lidar_fig} right), FBM (Table. \ref{tab:perclass_fmb_iou}), and MFNet (Table. \ref{tab:perclass_mfnet_iou}) datasets. Our proposed method, StitchFusion+FFMs, demonstrates balanced and robust performance across categories, achieving high mIOU scores. This comprehensive analysis not only highlights the strengths and weaknesses of each method but also provides valuable insights into their applicability across different datasets and categories, guiding further improvements in segmentation techniques.
For instance, in the FMB dataset evaluation, our models, "StitchFusion" and "StitchFusion+FFMs," demonstrate superior performance across various classes. Notably, "StitchFusion" achieves an 88.5\% mIoU in the "Building" class, reflecting its proficiency in architectural segmentation. The incorporation of Feature Fusion Modules (FFMs) in "StitchFusion+FFMs" notably boosts the mIoU for the "Traffic Lamp" class to 52.0\%, from the base model's 38.7\%, highlighting the benefits of advanced feature integration. Both models excel in the "Traffic Sign" class, with mIoUs of 83.6\% and 80.4\%, respectively, underscoring their reliability in critical traffic element recognition. These results indicate the robustness of our models in RGB-infrared segmentation tasks.

\begin{table*}[ht]
\centering
\small
\caption{Per-class mIoU (\%) comparison on FMB dataset for RGB-infrared modalities of StitchFusion(MiT-B3).
Bold represents the first, and underline represents the second.}
\label{tab:perclass_fmb_iou}
\vspace{-10pt}
\begin{tabular}{llcccccccccc}
\shline
\textbf{Methods} & \textbf{Car} & \textbf{Person} & \textbf{Truck} & \textbf{T-Lamp} & \textbf{T-Sign} & \textbf{Building} & \textbf{Vegetation} & \textbf{Pole} & \textbf{mIoU(\%)} \\ \shline
GMNet   & 79.3 & 60.1 & 22.2 & 21.6 & 69.0 & 79.1 & 83.8 & 39.8 & 49.2 \\ 
LASNet   & 72.6 & 48.6 & 14.8 & 2.9 & 59.0 & 75.4 & 81.6 & 36.7 & 42.5 \\ 
EGFNet   & 77.4 & 63.0 & 17.1 & 25.2 & 66.6 & 77.2 & 83.5 & 41.5 & 47.3 \\ 
FEANet   & 73.9 & 60.7 & 32.3 & 13.5 & 55.6 & 79.4 & 81.2 & 36.8 & 46.6 \\ 
DIDFuse   & 77.7 & 64.4 & 28.8 & 29.2 & 64.4 & 78.4 & 82.4 & 41.8 & 50.6 \\ 
ReCoNet  & 75.9 & 65.8 & 14.9 & 34.7 & 66.6 & 79.2 & 81.3 & 44.9 & 50.9 \\ 
U2Fusion  & 76.6 & 61.9 & 14.4 & 28.3 & 68.9 & 78.8 & 82.2 & 42.2 & 47.9 \\ 
TarDAL  & 74.2 & 56.0 & 18.8 & 29.6 & 66.5 & 79.1 & 81.7 & 41.9 & 47.9 \\ 
SegMiF  & 78.3 & 65.4 & \textbf{47.3} & 43.1 & 74.8 & 82.0 & 85.0 & 49.8 & 54.8 \\ 
MMSFormer  & 82.6 & 69.8 & \underline{44.6} & \underline{45.2} & \underline{79.7} & 83.0 & \underline{87.3} & \underline{51.4} & 61.7 \\ \shline
StitchFusion &  \textbf{83.3} & \textbf{75.1} & 42.8 & 38.7 & 78.9 & \textbf{85.1} & \textbf{88.5} & \textbf{52.1} & \underline{63.3} \\
StitchFusion+FFMs & \underline{83.0} & \underline{73.0} & 42.6 & \textbf{52.0} & \textbf{80.4} & \underline{83.6} & \textbf{88.5} & 49.5 & \textbf{64.3}\\
\shline
\end{tabular}
\end{table*}

\begin{table*}[ht]
\centering
\small
\caption{Per-class results on MFNet dataset for RGB-Thermal segmentation using StitchFusion(MiT-B4). 
Bold represents the first, and underline represents the second.}
\vspace{-10pt}
\label{tab:perclass_mfnet_iou}
\begin{tabular}{lccccccccccc}
\shline
\textbf{Method} & \textbf{Unlabeled} & \textbf{Car} & \textbf{Person} & \textbf{Bike} & \textbf{Curve} & \textbf{Car Stop} & \textbf{Guardrail} & \textbf{Color Cone} & \textbf{Bump} & \textbf{mIoU (\%)} \\ \shline
MFNet  & 96.9 & 65.9 & 58.9 & 42.9 & 29.9 & 9.9 & 0.0 & 25.2 & 27.7 & 39.7 \\ 
SA-Gate  & 96.8 & 73.8 & 59.2 & 51.3 & 38.4 & 19.3 & 0.0 & 24.5 & 48.8 & 45.8 \\ 
DA-CNN  & 96.9 & 77.0 & 53.4 & 56.5 & 30.9 & 29.3 & 8.5 & 30.1 & 32.4 & 46.1 \\ 
ACNet & 96.7 & 79.4 & 64.7 & 52.7 & 32.9 & 28.4 & 0.8 & 16.9 & 44.4 & 46.3 \\ 
PSTNet & 97.0 & 76.8 & 52.6 & 55.3 & 29.6 & 25.1 & \textbf{15.1} & 39.4 & 45.0 & 48.4 \\ 
RTFNet & \textbf{98.5} & 87.4 & 70.3 & 62.7 & 45.3 & 29.7 & 2.0 & 29.1 & 55.7 & 53.2 \\ 
FuseSeg  & 97.6 & 80.7 & 61.0 & \textbf{66.4} & 44.8 & 22.7 & 6.4 & 46.9 & 47.9 & 54.5 \\ 
AFNet & 98.0 & 86.0 & 67.4 & 62.6 & 37.5 & 28.9 & 4.6 & 47.4 & 50.0 & 53.6 \\ 
ABMDRNet & 98.6 & 84.8 & 69.6 & 60.7 & 33.1 & \textbf{33.3} & 5.1 & 47.4 & 50.3 & 54.8 \\ 
FEANet & 97.3 & 87.8 & 71.1 & 61.1 & 46.5 & 22.1 & 9.5 & 49.3 & 46.4 & 55.3 \\ 
DHIFNet  & 97.7 & 87.7 & 67.1 & 63.4 & 39.5 & 42.4 & 9.5 & 49.3 & \underline{56.0} & 57.3 \\ 
GMNet & 97.5 & 86.5 & 73.1 & 61.7 & 41.4 & 19.3 & \underline{14.5} & 48.7 & 48.8 & 57.3 \\ \shline
StitchFusion  & \underline{98.3} & \underline{89.5} & \textbf{75.1} & 65.6 & 47.9 & \underline{32.8} & 0.5 & \textbf{57.2} & 53.4 & \underline{57.8}\\
SF+FFMs  & \underline{98.3} & \textbf{89.7} & \textbf{75.1} & \underline{66.1} & \underline{45.7} & 28.3 & 8.1 & \underline{52.9} & \textbf{57.0} & \textbf{57.9}\\
\shline
\end{tabular}
\end{table*}
\input{table/07_Analysis_param_backbone_density} 

\section{Additional Parameter Number of StitchFusion For Different Backbone and Density.}
\label{sec:addition_para_app}
Table~\ref{tab:02_Param_Effi} demonstrates the scalability of StitchFusion's parameter addition across different backbones and input densities. The additional parameters increase proportionally with the number of modalities and backbone complexity, showcasing linear growth at each stage. For instance, when the backbone changes from MiT-B2 to ConvNext-Large with four input modalities, the total additional parameters increase from 0.908M to 5.868M, indicating that the design adapts efficiently to more complex configurations. Furthermore, the mean parameter addition per stage reflects a bounded increase, ensuring that StitchFusion maintains computational efficiency while scaling to diverse modalities and backbone architectures. This validates its adaptability for multimodal tasks without incurring excessive parameter overhead.

\section{Semantic Segmentation Visulization}
\label{sec:vis_app}
\subsubsection{t-SNE visualization.} 
\begin{figure}[ht]
\centering
\includegraphics[width=\columnwidth]{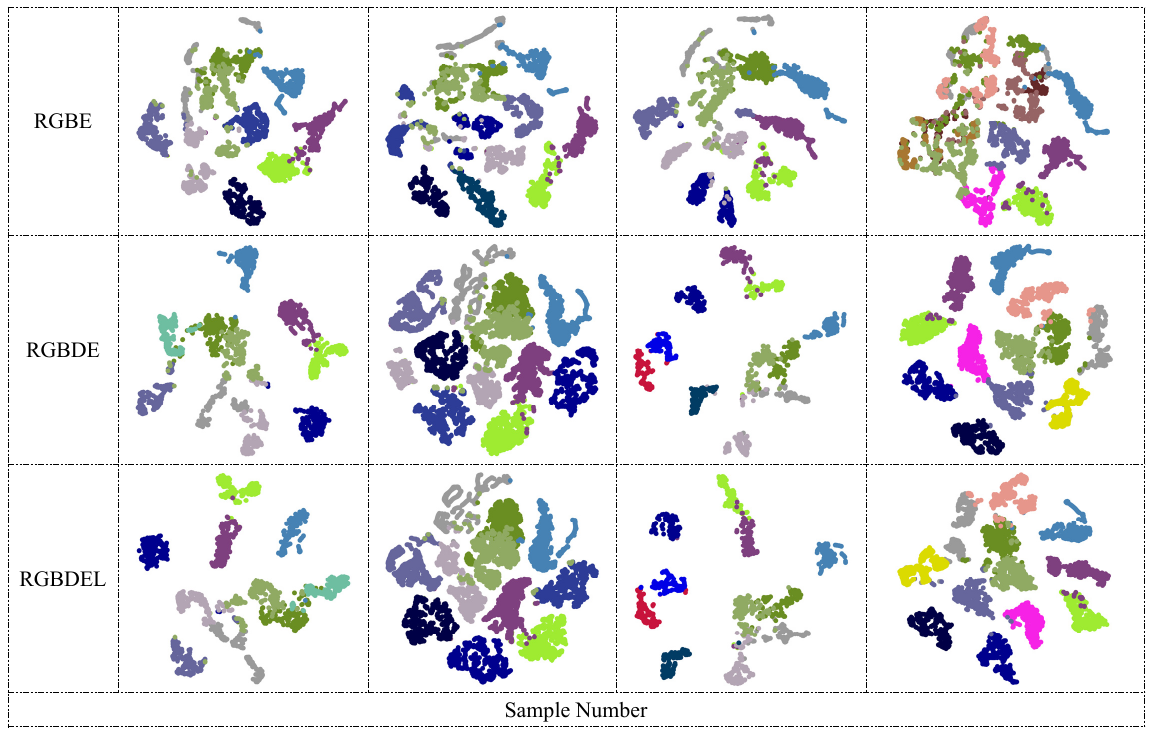}
\caption{t-SNE Visualization On DeLiVER Dataset.}
\label{fig:fig_tsne}
\end{figure}

The t-SNE visualization in Fig. \ref{fig:fig_tsne} demonstrates the clustering of features extracted by StitchFusion on the DeLiVER dataset. As the number of modalities increases from RGB to RGBE, RGBDE, and RGBDEL, the clusters become more distinct and well-separated, indicating enhanced feature differentiation. This suggests that incorporating additional modalities improves the model’s ability to learn more discriminative features.

\subsection{Supplementary Visualization of StitchFusion}

\begin{figure}[ht]
\centering
\includegraphics[width=\linewidth]{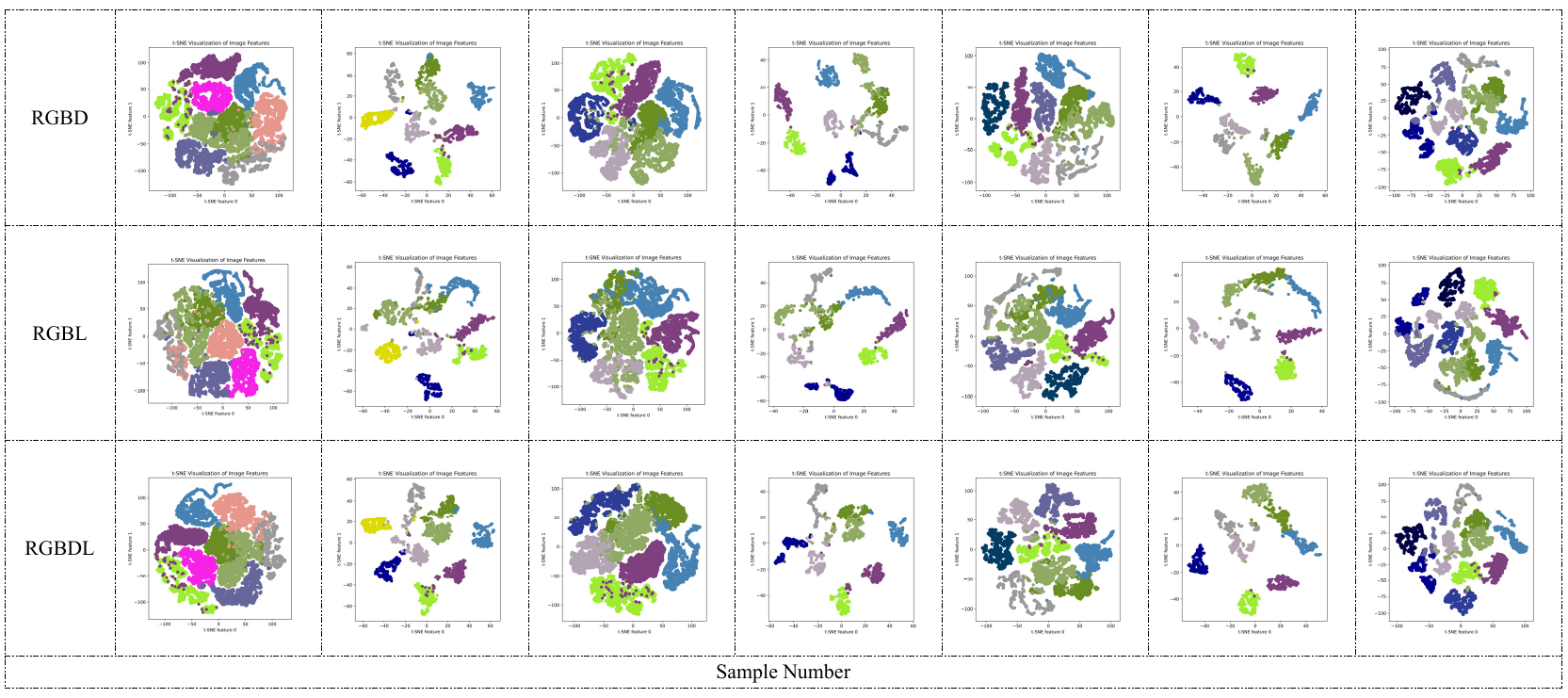}
\vspace{-10pt}
\caption{Supplementary t-SNE Visualization of StitchFusion (MiT-B2) On DeLiVER Dataset.}
\label{fig:fig_tsne2}
\end{figure}

\begin{figure}[ht]
\centering
\includegraphics[width=\linewidth]{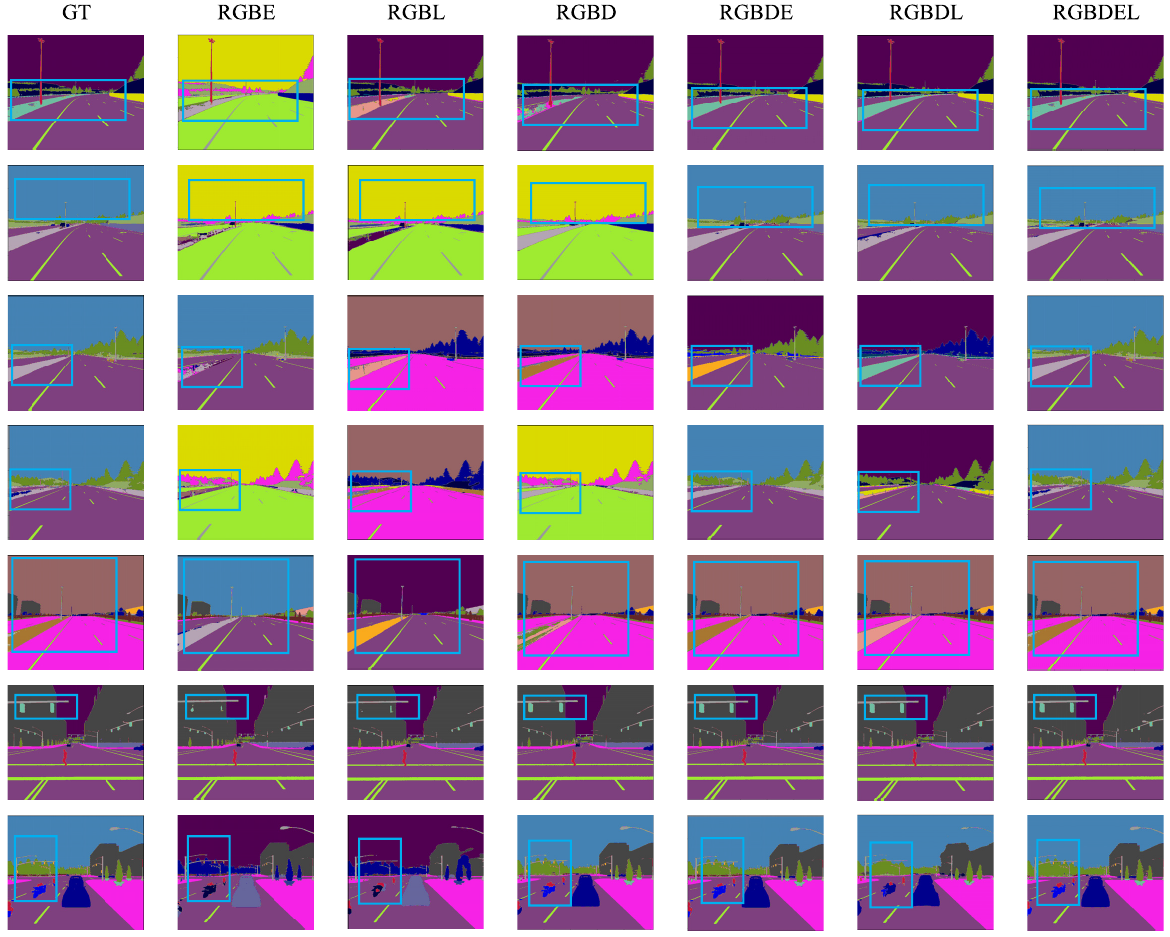}
\caption{Supplementary Visulization of StitchFusion (MiT-B2) On DeLiVER Dataset.}
\label{fig:visulizition_deliver_1}
\end{figure}

In this appendix, we provide a comprehensive set of visualizations and analyses to elucidate the performance and capabilities of the StitchFusion method.  To further underscore the method's feature extraction ability, Fig. \ref{fig:fig_tsne2} employs t-SNE to project the high-dimensional feature representations onto a two-dimensional plane. The distinct clusters in the t-SNE plots indicate the method's ability to capture and separate the nuances within the data. Fig. \ref{fig:visulizition_deliver_1} offers a detailed visualization of the segmentation results on the DeLiVER dataset, showcasing the method's precision in delineating boundaries and preserving fine details. 

\begin{figure}[ht]
\centering
\includegraphics[width=\linewidth]{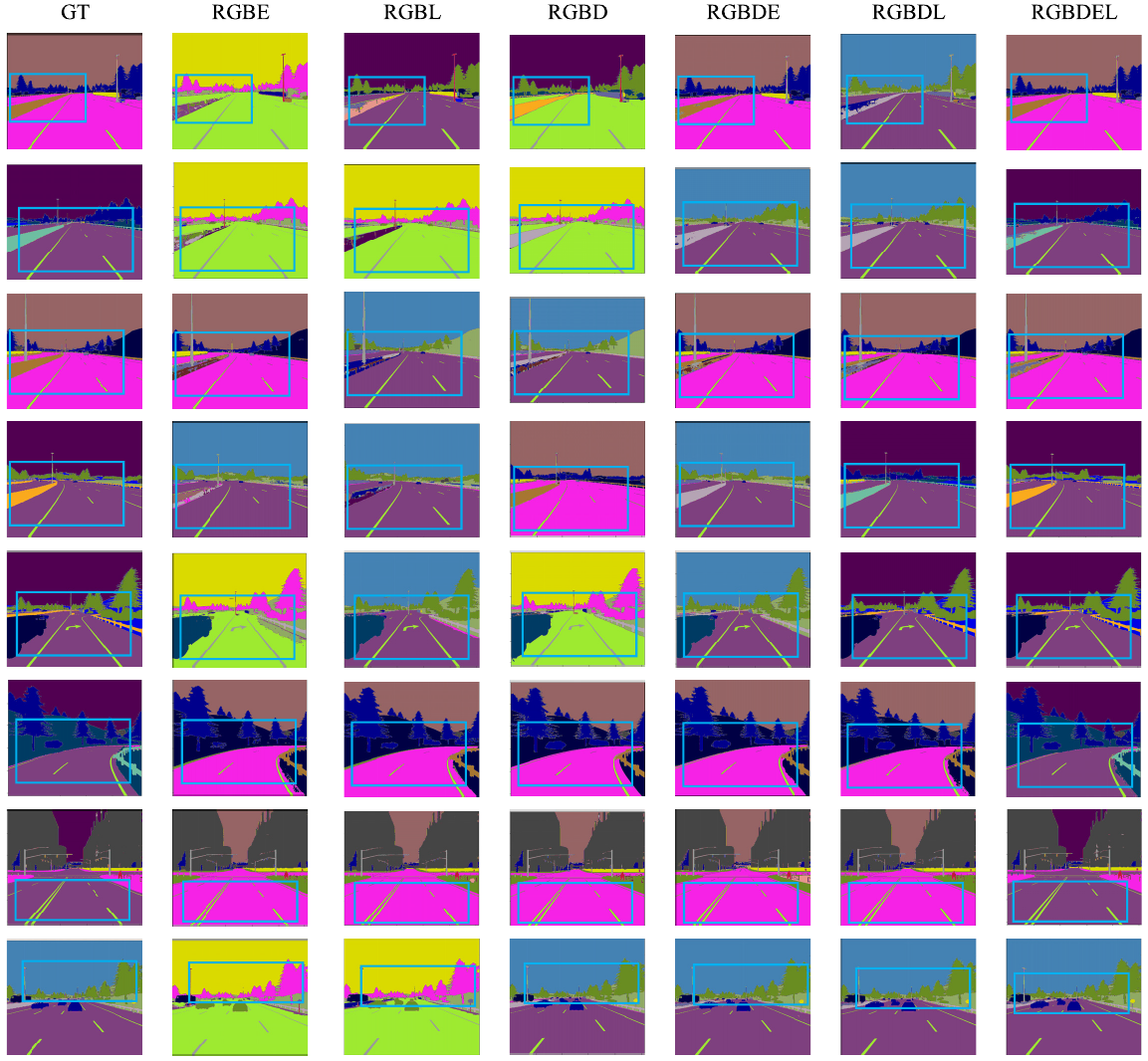}
\caption{Supplementary Visualization of StitchFusion (MiT-B2) On DeLiVER Dataset.}
\label{fig:visulizition_deliver_2}
\end{figure}

\begin{figure}[ht]
\centering
\includegraphics[width=0.9\linewidth]{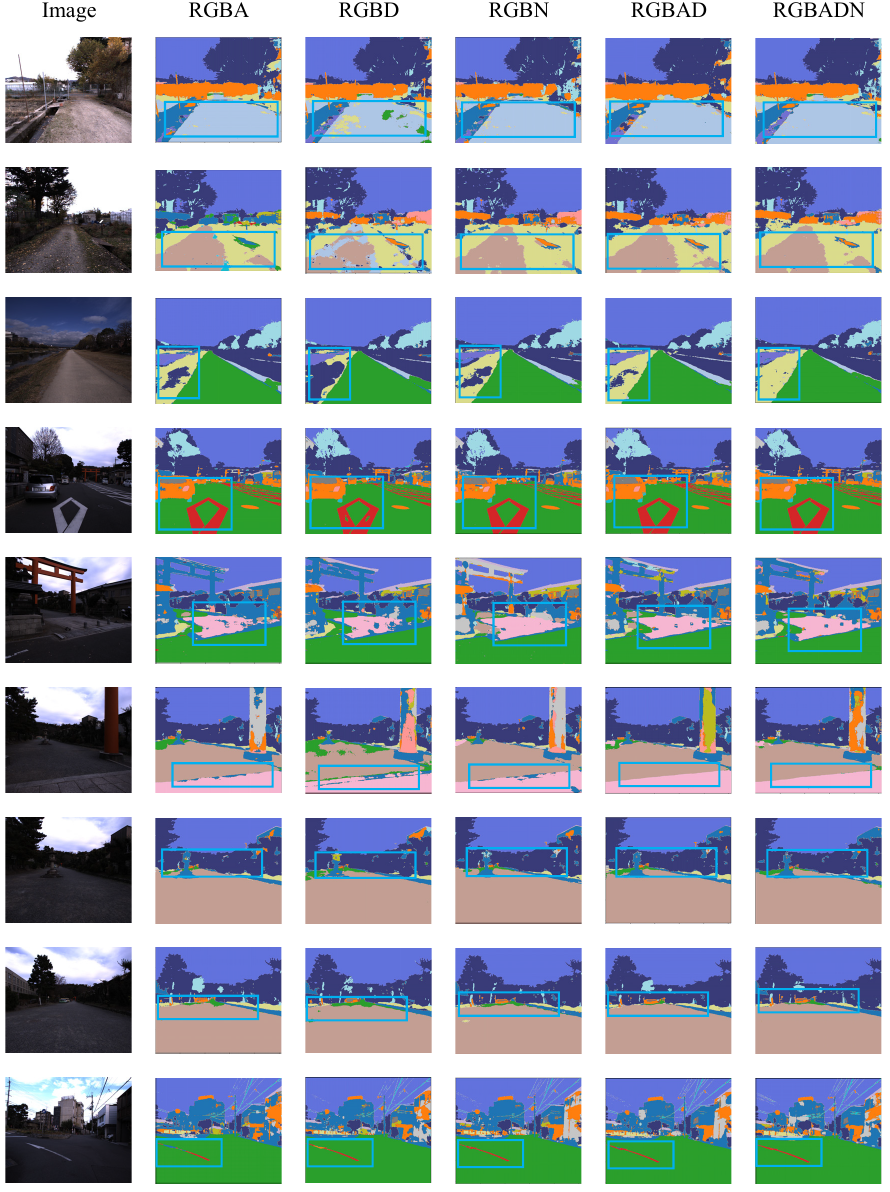}
\caption{Supplementary Visualization of StitchFusion (MiT-B4) On Mcubes Dataset.}
\label{fig:Visulization_mcubes_2}
\end{figure}

Fig. \ref{fig:visulizition_deliver_2} provides more visualizations of StitchFusion's segmentation outcomes on the DeLiVER dataset, offering a broader perspective on the model's performance. 
Finally, Fig. \ref{fig:Visulization_mcubes_2} extends the qualitative assessment of the Mcubes dataset by presenting additional segmentation results, thereby demonstrating the model's adaptability across varied data environments.
These additional images may include comparative analyses with other methods or highlight the method's efficacy in handling complex or challenging segments within the dataset.

%% file: table/07_Analysis_param_backbone_density.tex
\begin{table*}[htbp]
    \centering
    \caption{Additional Parameter Efficiency, We set hidden dimension $r$ equal to 4.}
    \vspace{-10pt}
    \resizebox{0.8\linewidth}{!}{
    \begin{tabular}{cl|cccccccc}
        \shline
        \multicolumn{2}{c}{\textbf{Backbone and Modals}} & \multicolumn{8}{c}{\textbf{Additional Params. (M)} } \\
        \shline
        \textbf{Modals} &\textbf{Backbone} & \textbf{Stage 0} & \textbf{Stage 1} & \textbf{Stage 2} & \textbf{Stage 3} & \textbf{Stage 0-1} & \textbf{Stage 2-3} & \textbf{Stage-All} & \textbf{Mean}\\
        \shline
        \multirow{7}{*}{2} & MIT-B2  & 0.007 & 0.019 & 0.070 & 0.056 & 0.026 & 0.125 & 0.151 & 0.065 \\
        & MIT-B4  & 0.007 & 0.038 & 0.314 & 0.056 & 0.045 & 0.369 & 0.414 & 0.177 \\
        & Swin-Tiny & 0.007 & 0.014 & 0.084 & 0.055 & 0.021 & 0.139 & 0.160 & 0.069 \\
        & Swin-Small & 0.007 & 0.014 & 0.251 & 0.055 & 0.021 & 0.306 & 0.327 & 0.140 \\
        & Swin-Large  & 0.014 & 0.028 & 0.499 & 0.111 & 0.042 & 0.610 & 0.652 & 0.279 \\
        & ConvNext-Base  & 0.014 & 0.028 & 0.500 & 0.111 & 0.042 & 0.611 & 0.653 & 0.280 \\
        & ConvNext-Large  & 0.021 & 0.042 & 0.749 & 0.166 & 0.063 & 0.915 & 0.978 & 0.419 \\
        \shline
        \multirow{7}{*}{3} & MIT-B2 & 0.022 & 0.056 & 0.209 & 0.167 & 0.078 & 0.376 & 0.454 & 0.195 \\
        & MIT-B4  & 0.022 & 0.113 & 0.941 & 0.167 & 0.134 & 1.108 & 1.242 & 0.532 \\
        & Swin-Tiny  & 0.021 & 0.042 & 0.251 & 0.166 & 0.063 & 0.417 & 0.480 & 0.206 \\
        & Swin-Small & 0.021 & 0.042 & 0.752 & 0.166 & 0.063 & 0.918 & 0.982 & 0.421 \\
        & Swin-Large  & 0.042 & 0.084 & 1.498 & 0.332 & 0.126 & 1.831 & 1.956 & 0.838 \\
        & ConvNext-Base  & 0.042 & 0.084 & 1.501 & 0.333 & 0.126 & 1.833 & 1.960 & 0.840 \\
        & ConvNext-Large  & 0.063 & 0.125 & 2.247 & 0.499 & 0.188 & 2.746 & 2.934 & 1.257 \\
        \shline
        \multirow{7}{*}{4} & MIT-B2 & 0.043 & 0.113 & 0.418 & 0.334 & 0.156 & 0.752 & 0.908 & 0.389 \\
        & MIT-B4  & 0.043 & 0.226 & 1.882 & 0.334 & 0.269 & 2.215 & 2.484 & 1.065 \\
        & Swin-Tiny  & 0.043 & 0.084 & 0.501 & 0.333 & 0.127 & 0.834 & 0.961 & 0.412 \\
        & Swin-Small & 0.043 & 0.084 & 1.503 & 0.333 & 0.127 & 1.836 & 1.963 & 0.841 \\
        & Swin-Large & 0.084 & 0.167 & 2.996 & 0.665 & 0.251 & 3.661 & 3.912 & 1.677 \\
        & ConvNext-Base  & 0.085 & 0.168 & 0.168 & 0.665 & 0.252 & 0.833 & 1.085 & 0.465 \\
        & ConvNext-Large & 0.126 & 0.251 & 4.495 & 0.997 & 0.377 & 5.492 & 5.868 & 2.515 \\
        \shline
    \end{tabular}}
    \label{tab:02_Param_Effi}
\end{table*}